%% file: kdd2027_main.tex
\documentclass[sigconf]{acmart}
\AtBeginDocument{%
  }
\usepackage{amsfonts,amsmath}
\usepackage{array,booktabs,tabularx}
\usepackage{enumitem}
\usepackage{placeins}
\usepackage{eso-pic}
\usepackage{hyperref}
\usepackage{url}
\usepackage{xcolor}
\definecolor{mycitecolor}{RGB}{28, 88, 140}
\definecolor{mylinkcolor}{RGB}{23, 90, 76}
\newcolumntype{Y}{>{\raggedright\arraybackslash}X}
\hypersetup{
    colorlinks=true,
    citecolor=mycitecolor, 
    linkcolor=mylinkcolor, 
    urlcolor=mycitecolor
}

\setcopyright{none}
\renewcommand\footnotetextcopyrightpermission[1]{}

\newcommand{\emailmark}[1]{\textsuperscript{\normalfont\scriptsize #1}}
\newcommand{\correspondingmark}{\textsuperscript{\normalfont\scriptsize *}}
\newcommand{\firstpageauthoremails}{%
  \AddToShipoutPictureFG*{%
    \AtPageLowerLeft{%
      \raisebox{0.28in}[0pt][0pt]{%
        \hspace*{\dimexpr 1in+\oddsidemargin\relax}%
        \parbox[b]{\columnwidth}{%
          \scriptsize\raggedright
          \rule{1.25in}{0.4pt}\\[-0.2ex]
          \textit{Author emails (numbers match the author list):}\\[-0.1ex]
          \emailmark{1}\texttt{jiatong.li@connect.polyu.hk};
          \emailmark{2}\texttt{weng-yu.zhang@connect.polyu.hk};\\[-0.1ex]
          \emailmark{3}\texttt{wangweida@pjlab.org.cn};
          \emailmark{4}\texttt{yuxuan.ren@nus.edu.sg};\\[-0.1ex]
          \emailmark{5}\texttt{captain.130@sjtu.edu.cn};
          \emailmark{6}\texttt{25047313g@connect.polyu.hk};\\[-0.1ex]
          \emailmark{7}\texttt{liyuqiang@pjlab.org.cn};
          \emailmark{8}\texttt{ybian@nus.edu.sg};\\[-0.1ex]
          \emailmark{9}\correspondingmark\texttt{changmeng.zheng@polyu.edu.hk};
          \emailmark{10}\correspondingmark\texttt{x1wei@polyu.edu.hk};\\[-0.1ex]
          \emailmark{11}\texttt{csqli@comp.polyu.edu.hk}.\\[-0.1ex]
          \correspondingmark\textit{Corresponding authors: Changmeng Zheng and Xiaoyong Wei.}%
        }%
      }%
    }%
  }%
}

\author{Jiatong Li\emailmark{1}}
\affiliation{%
  \institution{The Hong Kong Polytechnic University}
  \city{Hong Kong}
  \country{China}
}

\author{Wengyu Zhang\emailmark{2}}
\affiliation{%
  \institution{The Hong Kong Polytechnic University}
  \city{Hong Kong}
  \country{China}
}

\author{Weida Wang\emailmark{3}}
\affiliation{%
  \institution{Shanghai AI Lab}
  \city{Shanghai}
  \country{China}
}

\author{Yuxuan Ren\emailmark{4}}
\affiliation{%
  \institution{National University of Singapore}
  \country{Singapore}
}

\author{Wei Liu\emailmark{5}}
\affiliation{%
  \institution{Shanghai Jiao Tong University}
  \city{Shanghai}
  \country{China}
}

\author{Chenyang Mao\emailmark{6}}
\affiliation{%
  \institution{The Hong Kong Polytechnic University}
  \city{Hong Kong}
  \country{China}
}

\author{Yuqiang Li\emailmark{7}}
\affiliation{%
  \institution{Shanghai AI Lab}
  \city{Shanghai}
  \country{China}
}

\author{Yatao Bian\emailmark{8}}
\affiliation{%
  \institution{National University of Singapore}
  \country{Singapore}
}

\author{Changmeng Zheng\emailmark{9}\correspondingmark}
\affiliation{%
  \institution{The Hong Kong Polytechnic University}
  \city{Hong Kong}
  \country{China}
}

\author{Xiaoyong Wei\emailmark{10}\correspondingmark}
\affiliation{%
  \institution{The Hong Kong Polytechnic University}
  \city{Hong Kong}
  \country{China}
}

\author{Qing Li\emailmark{11}}
\affiliation{%
  \institution{The Hong Kong Polytechnic University}
  \city{Hong Kong}
  \country{China}
}

\begin{document}

\title{Molecular LLM Agents: From Architectural Design to Scientific Autonomy}

\settopmatter{authorsperrow=4,printacmref=false}

\renewcommand{\shortauthors}{MolLLMAgent}

% Replace the ACM proceedings running text at the upper left with a neutral
% preprint label while retaining the existing right-side running information.
\makeatletter
\fancypagestyle{preprintpagestyle}{%
  \fancyhf{}%
  \renewcommand{\headrulewidth}{0pt}%
  \renewcommand{\footrulewidth}{0pt}%
  \fancyhead[LE,LO]{\@headfootfont Preprint}%
  \fancyhead[RE]{\@headfootfont\@shortauthors}%
  \fancyhead[RO]{\@headfootfont
    \acmConference@shortname, \acmConference@date, \acmConference@venue}%
}
\fancypagestyle{preprintfirstpagestyle}{%
  \fancyhf{}%
  \renewcommand{\headrulewidth}{0pt}%
  \renewcommand{\footrulewidth}{0pt}%
  \fancyhead[L]{\@headfootfont Preprint}%
}
\makeatother

\input{sections/00abs}




\firstpageauthoremails
\maketitle
\pagestyle{preprintpagestyle}

\input{sections/01introduction}

\input{sections/03framework-00_Overall}

\input{sections/03framework-05-Benchmarking}

\input{sections/04autonomy}

\input{sections/05safety_challenges}

\input{sections/06conclusion}

\bibliographystyle{ACM-Reference-Format}
\bibliography{sample-base}

% \appendix

% \input{sections/07appendix}

\end{document}

%% file: sections/00abs.tex
\begin{abstract}

Molecular science represents an important frontier for LLM-based agents.
Unlike general agents that mainly operate over natural language, code, or web environments, 
\emph{molecular LLM agents} must perceive, reason about, and act upon chemical objects across symbolic strings, molecular graphs, 3D conformations, spectra, simulations, and wet-lab measurements.
Their capabilities depend on chemically faithful molecular perception, an LLM-centered agent framework, domain-specific tool grounding, and computational or experimental feedback, in addition to planning and tool use.
This work develops a conceptual framework for molecular LLM agents from two complementary perspectives.
\textbf{First}, we introduce an architectural view of molecular-agent design, covering molecular representation and perception, the agent framework, domain-specific toolboxes, and learning and optimization.
\textbf{Second}, we propose a scientific autonomy ladder inspired by staged autonomy in engineering systems, categorizing agents into four levels: \textbf{L1} assistive or fixed workflows, \textbf{L2} adaptive computational agents, \textbf{L3} feedback-aware physical experiment agents, and \textbf{L4} scientific-agenda agents.
Together, these two perspectives establish a comprehensive framework for comparing existing molecular LLM agents, identifying missing capabilities and deployment risks, and guiding the design, evaluation, and deployment of future agents in molecular discovery workflows.

\end{abstract}

%% file: sections/01introduction.tex
\section{Introduction}

\begin{figure}[t]
\centering
\includegraphics[width=\linewidth]{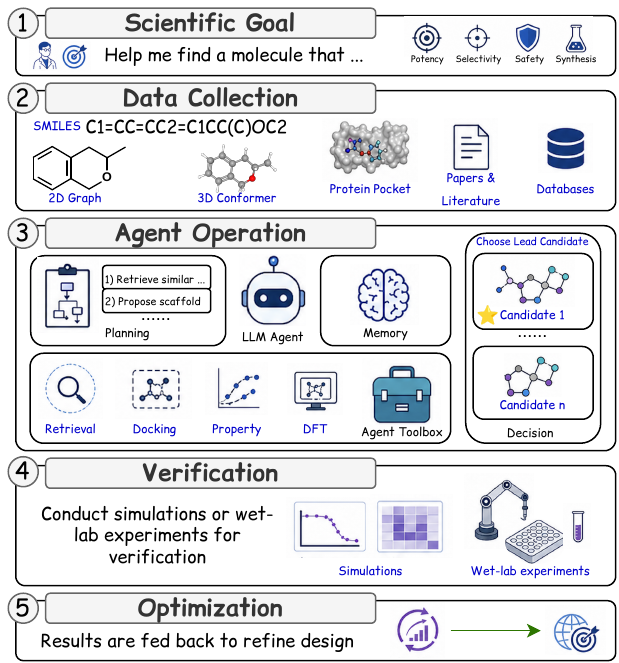}
\vskip -0.1in
\caption{End-to-end workflow of a molecular LLM agent. Starting from a scientific goal and task-specific constraints, the agent gathers multi-modal molecular evidence, plans and executes chemistry-tool operations, selects candidates, and verifies them through simulation or wet-lab experiments. Verification results are then fed back to refine the molecular design and the agent's subsequent actions.}
\Description{Five numbered stages connect a scientific goal, multi-modal data collection, agent planning and tool use, computational or wet-lab verification, and feedback-driven optimization.}
\label{fig:introduction}
\vskip -0.2in
\end{figure}

Molecular discovery asks how to turn a desired function into a real molecular entity \cite{li2026speak, li2024icma}. The target may involve biological activity, selectivity, toxicity, stability, or other physicochemical and functional properties~\cite{wu2018moleculenet,huang2021tdc}.
A useful molecule must also be chemically valid, compatible with multiple objectives, synthetically accessible, safe for its intended context, and supported by computational or experimental evidence~\cite{brown2019guacamol,polykovskiy2020moses}.
The search space is vast, evaluation is costly and uneven, and progress depends on coordinating the appropriate representation, model, tool, and validation signal.

Modern molecular AI has strengthened many components of this process.
Graph-based models have improved structure-property prediction~\cite{yang2019analyzing}, while graph generative models support molecular generation and optimization in graph space~\cite{jin2018junction}.
Chemical language models such as ChemBERTa~\cite{chithrananda2020chemberta} and MoLFormer~\cite{ross2022large} learn molecular representations from SMILES~\cite{weininger1988smiles}.
MolT5~\cite{edwards2022translation} and MolReGPT~\cite{li2023molregpt} further connect molecular structures with natural language for captioning, generation, and in-context learning. Together with reaction prediction, synthesis planning, and molecular optimization, these methods provide a rich stack of capabilities.
Most nevertheless remain specialized models or fixed mappings, such as molecule to property, text to molecule, or candidate to score.

The practical bottleneck is therefore shifting from solving an individual subtask to orchestrating subtasks into a coherent discovery workflow~\citep{li2026molvibench}.
A chemist must translate a natural-language objective into molecular objects, choose among strings, graphs, conformers, and spectra, retrieve prior evidence, invoke scientific tools, reject invalid candidates, interpret noisy outputs, and decide what should happen next. Conventional molecular AI leaves most of this orchestration to the user. 
A model may predict, generate, or rank, while the user still decides which evidence to trust, how to repair failures, and when to continue, stop, or proceed toward synthesis and measurement.

Large language model (LLM) agents offer a route from isolated models to action-oriented molecular systems. The LLM acts as a controller that interprets a goal, decomposes it into actions, invokes tools, observes feedback, and revises the plan. 
ReAct~\cite{yao2023react} made this control pattern explicit by interleaving reasoning with actions, while Toolformer~\cite{schick2023toolformer} demonstrated language-model tool invocation. 
In molecular discovery, an LLM-centered system may strengthen one subtask through reasoning \cite{li2025molr1,wang2026chem}, retrieval or verification \cite{li2024molreflect}, or coordinate a longer trajectory involving structure lookup, property calculation, docking, quantum chemistry, molecular dynamics, retrosynthesis, and laboratory interaction. The key change is that the model participates in making decisions, selecting actions and interpreting observations rather than stopping after a single prediction or generation step.

The molecular setting makes this control problem distinct from general web or software agents \cite{ning2025survey}. 
The agent must preserve chemical identity, connectivity, stereochemistry, geometry, units, and experimental conditions while moving among representations and tools. 
Its observations range from inexpensive heuristic scores to simulations, spectra, assays, and hardware logs, each with different uncertainty and cost. 
Errors can therefore propagate from a textual decision into an invalid calculation, an expensive simulation, or a physical experiment. Fluent reasoning alone is not sufficient; actions and feedback must remain chemically grounded and auditable.

We use \emph{molecular LLM agent} to denote an LLM-centered decision system that operates on molecular states, reasons over scientific goals, selects actions through chemistry tools or environments, and uses computational or experimental observations to produce, validate, or refine molecular outcomes.
Figure~\ref{fig:introduction} summarizes this feedback-driven workflow from goal specification and data collection to agent operation, verification, and design revision. 
ChemCrow~\cite{m2024augmenting} coordinates chemistry tools for structure lookup, property calculation, reaction prediction, and synthesis planning. 
Coscientist~\cite{boiko2023autonomous} extends the pattern toward laboratory-facing execution and feedback. 
These systems illustrate the move from task-specific molecular models to controllers over perception, action, tools, and evidence.

Despite this progress, the field lacks a shared design framework, autonomy roadmap, and governance boundary. 
Existing systems differ in molecular representation, tool interfaces, memory, feedback, optimization, and human oversight, yet are often discussed under the same agent label. This ambiguity leaves three questions unresolved. 
Which components should a molecular agent contain, and how should they interact? 
How could fixed workflows, computational loops, physical experimentation, and open-ended scientific agency be distinguished? 
And how could increasingly consequential actions be evaluated and governed? Without a common framework, it is difficult to compare systems, identify missing capabilities, or assess deployment risks.

We address these questions through two complementary perspectives.
\textbf{The architectural view} decomposes molecular agents into molecular representation and perception, an LLM-centered agent framework, domain-specific toolboxes, and learning and optimization. It treats evaluation, safety, and trustworthy deployment as cross-cutting requirements rather than properties of the LLM alone. 
Section~\ref{sec:evaluation_benchmarking} separately consolidates evaluation settings for component capabilities, executable workflows, research tasks, and closed-loop discovery.
\textbf{The autonomy view} classifies systems by the outermost feedback loop they demonstrably close: L1 assistive or fixed workflows, L2 adaptive computational agents, L3 feedback-aware physical workflows, and L4 scientific-agenda agents. Section~\ref{sec:autonomy_levels} provides the operational definitions and boundary cases.

Our contributions are as follows:
\begin{itemize}
    \item We conceptualize molecular LLM agents as action-oriented systems that connect molecular state, agent control, scientific tools, and feedback.

    \item We develop a dual-perspective framework that links architectural design to an evidence-based L1 to L4 scientific-autonomy ladder.

    \item We use the framework to compare existing systems, identify capability and evaluation gaps, and characterize deployment and governance risks.
\end{itemize}

%% file: sections/03framework-00_Overall.tex
\section{Architecture Design of Molecular Agents}
\label{sec:architecture}

\begin{figure*}[t]
\centering
\includegraphics[width=\linewidth]{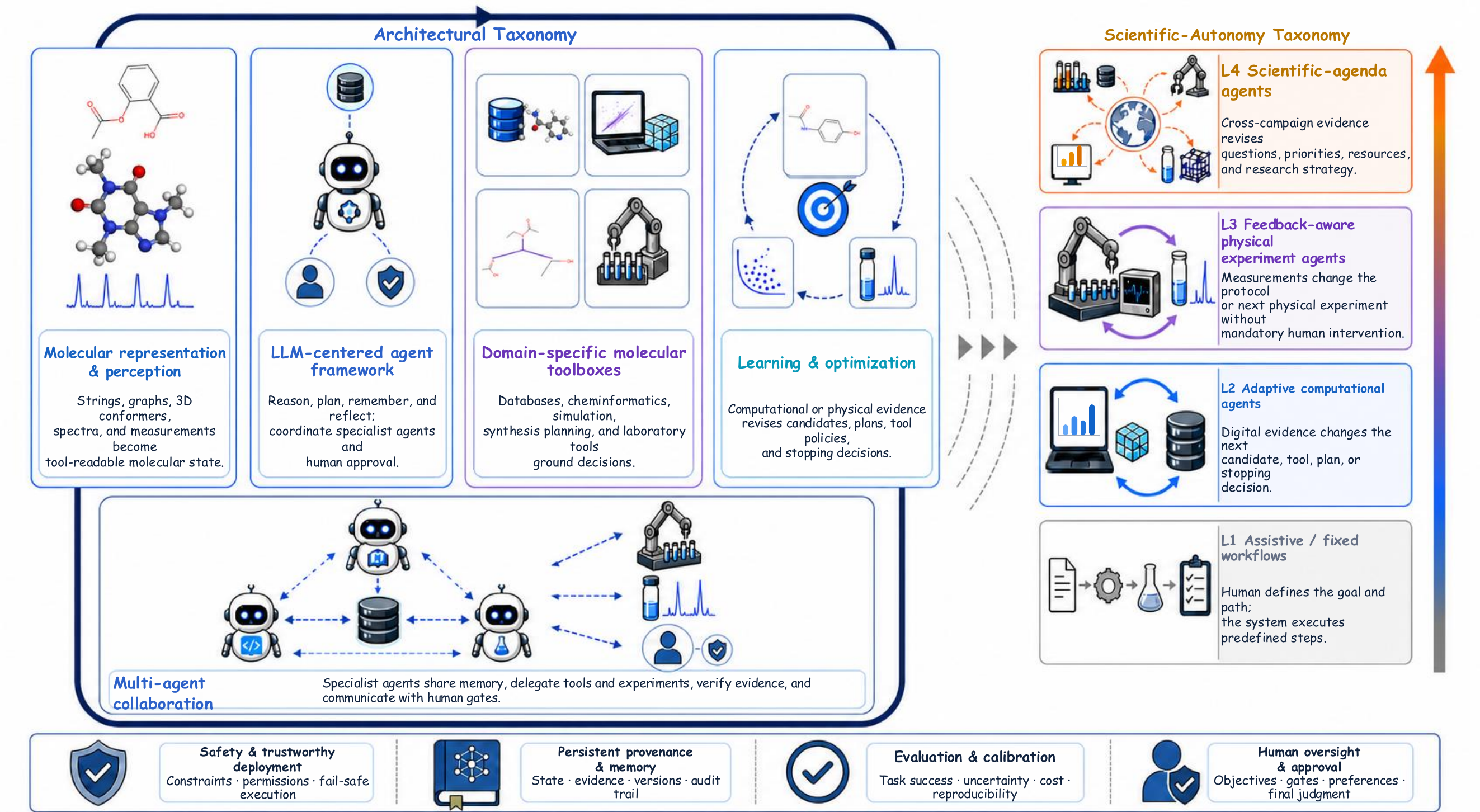}
\caption{\textbf{Two complementary taxonomies for molecular LLM agents.}
The architectural taxonomy (left) organizes agent systems around molecular representation and perception, an LLM-centered agent framework, domain-specific molecular toolboxes, and learning and optimization, together with multi-agent collaboration and cross-cutting deployment requirements.
The scientific-autonomy taxonomy (right) classifies systems by the outermost feedback loop they can reliably close without mandatory human intervention, ranging from L1 assistive or fixed workflows to L4 scientific-agenda agents.}
\Description{The left side shows four architectural components, with agent collaboration and safety surrounding a molecular LLM agent. The right side shows four increasing autonomy levels from assistive or fixed workflows to scientific-agenda discovery.}
\label{fig:architecture}
\end{figure*}

We organize a molecular LLM agent into four interacting components, as shown in Figure~\ref{fig:architecture}.
The \emph{perception layer} represents molecular objects and routes them among strings, graphs, geometries, images, and structured records. The \emph{agent framework} reasons over scientific goals, plans actions, maintains memory, reflects on feedback, and may coordinate specialized agents. The \emph{toolbox} grounds these decisions in databases, cheminformatics, simulation, synthesis, and experimental interfaces. Finally, \emph{reflection, learning, and optimization} convert computational or physical feedback into revisions of molecular candidates, plans, or reusable policies. These components form a closed chain from molecular state to decision, action, observation, and optimization.

\input{sections/03framework-01_Perception}
\input{sections/03framework-02_Agent_Design}
\input{sections/03framework-03-Tool}
\input{sections/03framework-04-Optimization}

%% file: sections/03framework-01_Perception.tex
\subsection{Molecular Representation and Perception}
\label{sec:molecular-perception}

Perception is the interface through which a molecular agent turns a chemical object into a state that an LLM-centered controller can read, edit, verify, and pass to tools.
This interface is more constrained than ordinary text perception: a molecule has atom identities, bond orders, aromaticity, charge, stereochemistry, conformers, electronic effects, and task-dependent physical context.
Consequently, the agent does not perceive ``the molecule itself'', but a representation of it.
The choice of representation influences which chemical facts are explicit, which facts must be inferred, and which actions are comparatively easy or brittle; these effects also depend on the model, task, and validation tools surrounding the representation.

\subsubsection{Representation Substrates: Strings, Graphs, Geometry, Images, and Structured Text}

The most common entry point is a 1D chemical line notation.
SMILES serializes a molecular graph into an ASCII string and remains attractive because it is compact, parser-friendly, and directly compatible with sequence models~\cite{weininger1988smiles}.
InChI offers another standardized identifier oriented toward chemical databases and interoperability~\cite{heller2015inchi}.
For LLM agents, these strings are convenient action interfaces: a controller can generate a candidate SMILES, call a toolkit to parse it, compute descriptors, search a database, or send it to a retrosynthesis service.
Their weakness is that the graph is implicit in a traversal.
Branches, ring closures, aromaticity, and stereochemical marks must be reconstructed from the sequence, and different valid strings can describe the same molecule.
This is acceptable for fast screening and tool calls, but fragile for tasks that require exact topology editing \cite{2026-molecular-generalization} or long-range structural reasoning.

SELFIES changes the failure mode by defining a robust string representation in which every SELFIES string maps to a chemically valid molecule under the representation's constraints~\cite{krenn2020selfies}.
This makes SELFIES especially useful when an agent repeatedly samples, mutates, or optimizes molecular strings, because local generation errors are less likely to collapse into unparsable outputs.
The guarantee, however, is a validity guarantee rather than a guarantee of stability, synthesizability, or task quality.
Fragment-level variants move the perceptual unit closer to how chemists design molecules.
Group SELFIES extends SELFIES with group tokens for common functional groups or substructures~\cite{cheng2023group}, while SAFE represents a molecule as an unordered sequence of connected fragment blocks that remains compatible with SMILES parsers~\cite{noutahi2023safe}.
MolLingo and mCLM continue this direction by using molecule-native or synthesis-friendly building blocks for LLM-powered agents and chemical language models~\cite{nguyen2026mollingo,edwards2025mclm}.
These representations are less character-centric and more action-centric: the agent can operate on scaffolds, linkers, substituents, motifs, or functional modules instead of only on individual symbols.

Two-dimensional molecular graphs expose the topology directly.
Atoms become nodes, bonds become edges, and node or edge features encode element type, valence, charge, aromaticity, bond order, and stereochemistry.
This makes graph encoders natural for property prediction and structure-aware generation; directed message passing is a representative example of this family~\cite{yang2019analyzing}.
The difficulty is that a graph is not natively an autoregressive text sequence.
An LLM agent therefore needs either a graph encoder whose continuous output is projected into the language model, or a discrete graph-to-text interface that turns nodes and bonds into tokens.
Recent structured molecular languages pursue the latter route.
MolJSON encodes atoms and bonds in a JSON schema designed for LLM reasoning~\cite{runcie2026moljson}, and MoleCode uses explicit node-edge-subgraph primitives so that connectivity and stereochemical information are visible in the context window~\cite{liu2026molecode}.
The broader lesson from these works is that structure-sensitive reasoning improves when the relational structure is not hidden inside a linearized string.

Three-dimensional representations add geometry.
Coordinates, distance matrices, conformer ensembles, protein pockets, docking poses, or crystal structures are necessary when the relevant property depends on spatial arrangement rather than only connectivity.
SchNet showed how neural models can learn from interatomic distances for quantum interactions~\cite{schutt2017schnet}, and GraphMVP aligns 2D graph representations with 3D geometry to transfer conformational information into molecular graph encoders~\cite{liu2022pretraining}.
For agents, 3D perception is expensive but often decisive: docking, molecular dynamics, transition-state reasoning, binding-pose inspection, and materials simulation require coordinates and physical units that a line notation does not provide.
The practical problem is that a molecule may have many low-energy conformers, and downstream decisions can be sensitive to how those conformers are generated, ranked, and passed between tools.

Images provide a fourth perceptual channel \cite{li2025chemvlm}.
A rendered skeletal formula is close to what chemists see in papers, patents, and lab notebooks.
MolScribe treats molecular structure recognition as image-to-graph generation~\cite{qian2023molscribe}, and MolSight studies progressive visual pretraining over molecular diagrams for property prediction~\cite{baranwal2026molsight}.
Image perception is valuable when the agent must read literature or multimodal records, but it also introduces optical recognition errors: atom labels, wedge bonds, charges, and crowded ring systems must be converted back into a valid graph before most tools can use them.

Experimental observations form a further perceptual substrate once an agent
interacts with instruments or laboratory records. Spectra, chromatograms,
assay tables, images, time series, protocol events, and robot logs describe a
molecular system under particular conditions rather than molecular identity
alone. They therefore require the agent to preserve units, acquisition
settings, sample identifiers, uncertainty, and provenance together with the
measured values. Coscientist interprets UV--Vis measurements inside a physical
task~\cite{boiko2023autonomous}; LLM-RDF uses reaction yields and spectral
analysis in an end-to-end synthesis-development workflow~\cite{ruan2024automatic};
and ORGANA combines visual feedback with experiment execution and
reporting~\cite{darvish2025organa}. These systems illustrate why experimental
records should be represented as state-bearing observations, not flattened
into unqualified text.

\begin{table*}[t]
\centering
\scriptsize
\setlength{\tabcolsep}{4pt}
\renewcommand{\arraystretch}{1.16}
\caption{Common molecular and experimental perceptual substrates for LLM-based molecular agents.}
\label{tab:perception-substrates}
\begin{tabularx}{\textwidth}{@{}p{0.16\textwidth}p{0.25\textwidth}p{0.27\textwidth}X@{}}
\toprule
\textbf{Substrate} &
\textbf{Typical form} &
\textbf{What it makes explicit} &
\textbf{Agent-level use and main risk} \\
\midrule
Line notation &
SMILES, InChI &
Compact identity and parser-compatible connectivity &
Fast generation, search, and tool calls; topology and stereochemistry must be recovered from a brittle sequence. \\
Robust or fragment string &
SELFIES, Group SELFIES, SAFE, molecule-native fragments &
Validity constraints or chemically meaningful blocks &
Useful for repeated editing and optimization; validity does not ensure synthesizability or property quality. \\
Graph or structured text &
Molecular graph, MolJSON, MoleCode &
Atoms, bonds, local topology, explicit node and edge identities &
Better for exact edits and graph reasoning; requires graph encoders or verbose structured contexts. \\
3D geometry &
Conformers, coordinates, pockets, docking poses &
Distances, angles, chirality, spatial contacts, physical state &
Necessary for binding, quantum, and simulation tasks; expensive and conformer-sensitive. \\
Image &
Skeletal diagram or scanned chemical figure &
Human-readable structural drawing and visual stereochemistry cues &
Connects agents to papers and notebooks; must be recognized into graph or string form before most tools can act. \\
Experimental record &
Spectra, chromatograms, assays, time series, protocol events, robot logs &
Measured response, conditions, units, acquisition context, and execution state &
Grounds interpretation and laboratory control; sample mismatch, lost metadata, or overconfident peak assignment can corrupt later decisions. \\
\bottomrule
\end{tabularx}
\end{table*}

\subsubsection{Tokenization: The Agent's Molecular Action Interface}
\label{sec:tokenization}

Once a molecule is written as text or structured text, the LLM still sees only token IDs.
For LLM-based molecular agents, tokenization is not a neutral preprocessing step: it defines the granularity at which the agent can act on molecules, the way errors propagate through tool chains, and whether the agent can later read and reflect on its own molecular edits.
The question is distinct from tokenization for specialized chemical language models such as ChemBERTa~\cite{chithrananda2020chemberta}, MoLFormer~\cite{ross2022large}, or SELFormer~\cite{yuksel2023selformer}, which train dedicated encoders on domain-specific corpora and can choose tokenizers optimized for their own architecture.
An LLM agent, by contrast, typically operates through a general-purpose language model whose tokenizer is fixed at pretraining.
Molecular strings must pass through this given tokenizer and remain actionable on the other side.
The agent-level consequences of this constraint vary across autonomy levels.

\paragraph{Character-level molecular tokens and the L1 baseline.}
A generic subword tokenizer (BPE, WordPiece, Unigram) trained on natural-language corpora can split multi-character elements such as \texttt{Cl} and \texttt{Br}, scatter bracketed atoms and stereochemical marks across several tokens, and merge chemical punctuation with atoms in arbitrary ways~\cite{temizer2022chemical}.
At L1, where a human or fixed workflow closes each consequential transition, this can be tolerable: the agent generates a SMILES candidate, and a human review or deterministic validation catches parsing errors before they reach a downstream tool.
Atom-wise tokenizers that preserve atom boundaries and bracketed atoms improve this baseline, and Smirk decomposes bracket atoms into glyph-level tokens to retain open-vocabulary coverage for organometallics, coordination complexes, and other chemistry that general tokenizers replace with undifferentiated unknown tokens~\cite{wadell2024smirk}.
Even at L1, coverage matters because agents may query patents, catalysts, materials, metal complexes, polymers, salts, and tool-produced intermediates whose chemistry falls outside the clean organic subset seen during pretraining.

\paragraph{Fragment-level tokens and L2 iteration efficiency.}
When the agent moves to L2 and autonomously iterates through cycles of proposal, evaluation, and revision, tokenization begins to shape the optimization dynamics.
A character-level action edits one atom symbol; replacing a substituent on a scaffold may require five to eight sequential edits.
Fragment-level tokenizers, including Group SELFIES~\cite{cheng2023group},
SAFE~\cite{noutahi2023safe}, and mCLM~\cite{edwards2025mclm}, elevate the
action unit to a chemically coherent motif such as a functional group,
scaffold, or linker. The same edit can then be expressed in a single step.
This can shorten the action sequence needed to express a scaffold-level edit
and may simplify search or credit assignment when the model and objective are
aligned with the chosen fragments. The realized gain in planning or
computational efficiency is system-dependent, however, and should not be
attributed to token granularity alone.
Fragment-level tokenization also changes the error profile.
At the character level, a mismatched ring-closure digit or unbalanced bracket
can cause an RDKit~\cite{landrum2016rdkit} parsing failure. Descriptor
computation and docking preparation then cannot proceed, so the workflow halts.
The agent must detect the failure, diagnose its source, and retry, consuming planning budget.
SELFIES's validity guarantee~\cite{krenn2020selfies} and fragment-level constraints reduce the space of syntactically malformed outputs. They can therefore limit one source of error propagation at the perception-to-tool interface, while leaving stability, synthesis feasibility, and task quality unresolved.
This benefit may be missed by language-modeling evaluation but can matter for L2
agents that generate and evaluate many candidates in a campaign. Tool
augmentation can otherwise degrade chemistry problem solving when inputs are
malformed~\cite{yu2025tooling,li2025chemhas}.

\paragraph{Structure-discretized tokens and the transparency-efficiency trade-off.}
A newer family of tokenizers discretizes molecular structure rather than strings.
UniMoT uses a molecule encoder, a causal Q-Former, and vector quantization to map molecular graph features into discrete tokens that can be added to an LLM vocabulary~\cite{zhang2024unimot}.
AtomDisc and VQ-Atom similarly use graph or geometry context to quantize atom-level chemical neighborhoods~\cite{zhang2025atomdisc,kimura2026vqatom}.
For agents, this can reduce the interface gap between continuous molecular encoders and symbolic action traces and compress complex substructural edits into single token predictions.
The cost is reflective opacity.
Agents reflect on their action trajectories by reviewing previous tool calls, diagnosing failures, and deciding what to retry~\cite{tang2025chemagent,li2026molecular}.
This reflection operates in the LLM's text space: character-level and fragment-level tokens are human-readable, so the agent and a human auditor can inspect which molecular edit was attempted and what went wrong.
Structure-discretized tokens are opaque. An agent may record ``I predicted code
437 $\to$ code 892'' without being able to express the corresponding structural
change unless it invokes an external decoder.
MolLingo~\cite{nguyen2026mollingo} proposes molecule-native representations that retain structural meaning, while LatentChem~\cite{ye2026latentchem} replaces verbose textual chain-of-thought with latent thinking; both illustrate the design tension.
At L2, where the agent must self-correct during iterative optimization, reflective opacity can be a significant bottleneck.

\subsubsection{Perception as a Routing and Interaction Layer}
\label{sec:perception-routing}

Current molecular agents rarely rely on a single representation throughout a task.
They route among representations according to the action being taken: generating a SMILES candidate, parsing it into a graph for editing, converting to 3D for docking, converting back to SMILES for a database query, and rendering an image for human review.
ChemCrow couples an LLM controller to chemistry tools so that natural-language requests can be translated into such operations~\cite{m2024augmenting}, and ChemAgent adds memory and tool-use policies so that the perceived state includes prior tool outputs, evidence, and task history~\cite{tang2025chemagent}.
Routing begins with entity resolution. A user-supplied name, CAS number, or
drawn structure is mapped to a canonical molecular representation through
name-to-structure conversion with resources such as OPSIN~\cite{lowe2011opsin}
or PubChem~\cite{kim2023pubchem}, followed by protonation-state assignment and
tautomer selection. Routing continues through every later representation
transition.
At L1, a human or fixed workflow directs each transition and catches errors before they propagate; the routing layer is effectively a bidirectional format translator.
As autonomy rises, routing must become an autonomous state-management system with two capabilities that are largely unaddressed in existing architectures: conversion fidelity and bidirectional tool-facing perception.

\paragraph{Conversion validation.}
Each representation transition can lose information: stereochemistry may not survive a SMILES-to-graph-to-SMILES round trip; a conformer generator may fail silently; a structured-text encoding may truncate features that exceed the context-window budget.
Without validation checkpoints at each conversion boundary, the agent may silently operate on a different molecule from the one it started with.
MolJSON~\cite{runcie2026moljson} and MoleCode~\cite{liu2026molecode} make
relational structure visible in the context window, which supports
conversion-aware perception. Most agent architectures still lack systematic
checks that atom count, bond count, stereochemistry, and charge survive each
transition.
At L2, where the agent controls the routing, these checkpoints become mandatory infrastructure; a single undetected loss in cycle~3 of a 10-cycle optimization corrupts all subsequent cycles.
At L3, conversion validation extends to physical signals: the perception layer must verify both structural preservation and correct parsing, compound assignment, and condition annotation for each experimental readout.

\paragraph{Output perception and feedback perception.}
Before invoking a tool, the agent must transform its internal state into the tool's expected input format: a 3D SDF file for docking, canonical SMILES for a database query, charge and spin multiplicity for quantum chemistry~\cite{zou2025agente}, or a valid graph with explicit hydrogens for retrosynthesis.
Many agent failures originate at this output-perception boundary: the LLM reasons correctly about which tool to call but constructs an invalid input.
ChemHAS~\cite{li2025chemhas} and TRACE~\cite{li2026molecular} address this through self-correction at the tool interface, but recognizing these as perception failures rather than reasoning failures enables more targeted evaluation.
In the reverse direction, a tool output such as a docking score, ADMET
prediction, DFT energy, or spectral match must return to the agent's reasoning
space with context. The agent must determine whether the output corresponds to
the intended molecule, quantify its uncertainty, and identify conflicts with
other signals.
Tooling-or-Not-Tooling~\cite{yu2025tooling} shows that tool augmentation can help or hurt depending on context. The perception layer should therefore assess the reliability of each feedback signal before incorporating it into the agent state.
At L3, feedback perception faces a distinct challenge: experimental signals are inherently ambiguous, and the perception layer must preserve this ambiguity rather than collapsing to a single interpretation, since premature disambiguation can produce false experimental conclusions.
At L4, feedback from different campaigns, tools, and scoring functions must be compared on a common perceptual basis, which requires unified state representations and provenance tracking across projects.

Thus, molecular perception should route among multiple representations, remain
aware of tokenizer effects, validate conversions explicitly, and support
bidirectional interaction with tools.
The downstream controller can plan reliably only when its perceptual state
preserves the chemical constraints needed by the next action. This requirement
scales from human-checked format translation at L1 to cross-campaign state
unification at L4.

%% file: sections/03framework-02_Agent_Design.tex
\subsection{Agent Framework}
\label{sec:agent-framework}

The agent framework is the LLM-centered controller between molecular representation and scientific action. It selects tools or subagents, interprets their outputs, and decides whether a trajectory should continue, revise, stop, or request approval. ChemCrow~\cite{m2024augmenting} and CACTUS~\cite{mcnaughton2024cactus} illustrate chemistry-tool control; MDCrow~\cite{campbell2026mdcrow} and LLaMP~\cite{chiang2025llamp} extend it to longer computational workflows; and Coscientist~\cite{boiko2023autonomous}, LLM-RDF~\cite{ruan2024automatic}, and Tippy~\cite{fehlis2025accelerating,fehlis2025technical} move toward physical or laboratory-facing decisions.

\subsubsection{Overview: The Molecular Controller}

A molecular controller is judged by the scientific objects and actions it can produce, not by fluent text alone. Its outputs must remain chemically valid, its decisions must be grounded in noisy or costly evidence, and its internal state must preserve the molecular representation needed by downstream tools.
Most systems implement an observe, plan, act, and revise loop through five faculties: reasoning, planning, memory, reflection and self-correction, and
multi-agent collaboration.

\begin{figure*}[t]
\centering
\includegraphics[width=0.98\textwidth]{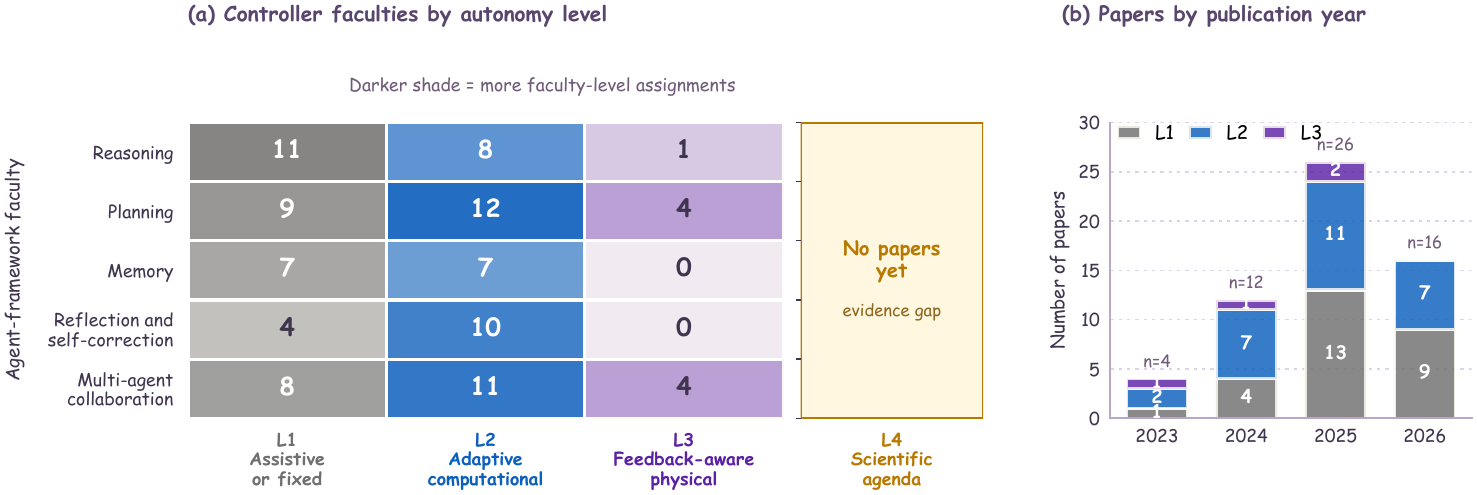}
\caption{Two views of the 58 surveyed papers by highest demonstrated autonomy level. The faculty heatmap counts 96 paper--faculty assignments, while the yearly stacked bars count each paper once; hatching marks the L4 evidence gap.}
\Description{Left: a faculty-by-autonomy heatmap for reasoning, planning, memory, reflection and self-correction, and multi-agent collaboration. Right: yearly stacked bars in which each paper is counted once at its highest demonstrated autonomy level. The L4 column is hatched because no surveyed paper qualifies.}
\label{fig:agent-framework-heatmap}
\end{figure*}

Figure~\ref{fig:agent-framework-heatmap} cross-tabulates papers by these faculties and by the highest demonstrated autonomy level under
Section~\ref{sec:autonomy_levels}. Each paper receives one global level that is reused across faculties: the heatmap counts it once per relevant faculty, whereas the yearly bars count it once overall.

\subsubsection{Reasoning}

Reasoning turns a molecular goal into hypotheses, intermediate conclusions, and candidate actions.
Early chemistry agents often express this reasoning as natural-language chain-of-thought over names, SMILES strings, reaction descriptions, or retrieved text.
Such reasoning is interpretable, but it is fragile: a small mistake in valence, stereochemistry, units, reaction feasibility, or tool-input syntax can invalidate a fluent explanation.
Chemistry-oriented LLMs therefore strengthen the reasoning substrate through domain training, memory, and molecular language modeling, as illustrated by ChemAgent~\cite{tang2025chemagent}, ether0~\cite{narayanan2026training}, mCLM~\cite{edwards2025mclm}, and comprehensive molecular design language models~\cite{yue2024unlocking}.

One direction replaces linear reasoning with search-augmented reasoning.
Molecular design is combinatorial, and many tasks require comparing multiple hypotheses rather than committing to one generated chain.
Monte Carlo Thought Search~\cite{sprueill2023monte} explores catalyst-design reasoning paths with tree search, while ChemReasoner~\cite{sprueill2024chemreasoner} searches an LLM's chemistry knowledge space and grounds the search with quantum chemical rewards.
CheMatAgent~\cite{wu2025chematagent} learns chemistry and materials tool-use policies through tree search based training, DrugMCTS~\cite{yang2025drugmcts} combines retrieval, multi-agent roles, and Monte Carlo Tree Search for drug repurposing, and Agents-on-a-Tree~\cite{zhang2026agents} coordinates pathwise molecular optimization.
A molecular reasoning branch can therefore be evaluated by docking, property prediction, adsorption energy, reaction barriers, quantum calculations, or other physical feedback instead of linguistic plausibility alone.

A second direction makes reasoning structured and executable.
Instead of producing only plain text, the agent emits tool calls, action sequences, code, protocol steps, or parameterized workflows that can be parsed, executed, checked, or replayed.
ChemActor~\cite{zhang2025chemactor} converts unstructured synthesis procedures into machine-executable chemical action sequences, MT-Mol~\cite{kim2025mt} decomposes molecular optimization into tool-based reasoning, and DrugPilot~\cite{li2025drugpilot} uses parameterized reasoning over multimodal drug-discovery information.
Coscientist~\cite{boiko2023autonomous}, El Agente~\cite{zou2025agente}, DrugAgent~\cite{liu2024drugagent}, and Chemist-X~\cite{chen2023chemist} expose reasoning through executable code, workflow traces, or computer-aided design interfaces.
%
% This trend also clarifies the current status of formal methods: in surveyed molecular agents, ``formal'' usually means structured, executable, or tool-checkable, not machine-verified in the Lean-style theorem-proving sense.

A third direction asks whether the agent should reason in text at all.
Text is inspectable, but it is lossy for graphs, conformers, stereochemical relations, and 3D interactions.
MolLingo proposes molecule-native representations for LLM scientific agents~\cite{nguyen2026mollingo}, and LatentChem replaces verbose textual chain-of-thought with latent thinking and a dynamic perception loop~\cite{ye2026latentchem}.
This creates a design trade-off: text supports human inspection, whereas molecule-native or latent states may preserve structural information that is hard to express in prose.
The boundary between reasoning and tool use is also adaptive rather than fixed, since tool augmentation can help some chemistry tasks and hurt others~\cite{yu2025tooling}.
Reasoning becomes autonomy-relevant only when its evidence changes a later
scientific action. ChemNavigator~\cite{peivaste2026chemnavigator}, for example,
extracts design rules within a bounded computational campaign and is therefore
an L2 boundary case rather than evidence of cross-campaign L4 discovery.

\subsubsection{Planning}

Planning turns a reasoning outcome into an executable trajectory.
For molecular agents, a plan must specify how information moves across representations, tools, constraints, and feedback signals.
A goal may require literature retrieval, candidate generation, filtering, docking, retrosynthesis, simulation, protocol generation, or laboratory execution.
Thus, planning lengthens the agent's horizon from one-shot tool use to in-silico campaigns.

A common pattern is task decomposition and workflow orchestration.
M$^4$olGen~\cite{li2026m} studies multi-stage generation under precise multi-property constraints, while Prompt-to-Pill~\cite{vichentijevikj2026prompt}, PharmAgents~\cite{gao2025pharmagents}, MADD~\cite{solovev2025madd}, and FROGENT~\cite{pan2026frogentendtoendfullprocessdrug} organize broader drug-discovery pipelines into stages or agents.
These systems show that generation, scoring, synthesis analysis, and reporting work closely together: each stage changes the state the planner must pass to the next module.

Another pattern is interleaved reason-act planning.
Rather than drafting a complete plan once, the agent alternates between local reasoning, tool invocation, observation, and revision.
ChemCrow~\cite{m2024augmenting} and CACTUS~\cite{mcnaughton2024cactus} exemplify this mode for chemistry tasks, LLaMP~\cite{chiang2025llamp} extends it to materials retrieval and simulations, and MDCrow~\cite{campbell2026mdcrow} applies it to molecular-dynamics workflows.
Longer tasks motivate hierarchy: El Agente~\cite{zou2025agente}, MASTER~\cite{rothfarb2025hierarchical}, robotic ChemAgents~\cite{song2025multiagent}, and Tippy~\cite{fehlis2025accelerating,fehlis2025technical} separate high-level scientific intent from mid-level workflow control and low-level execution.

Planning also includes tool selection and action-policy design.
In molecular agents, choosing RDKit, docking, retrosynthesis, DFT, molecular dynamics, a database, or a verifier is already a scientific decision.
ChemHTS~\cite{li2025chemhts} studies hierarchical tool stacking, and
ChemHAS~\cite{li2025chemhas} improves chemistry-tool performance through agent
stacking. TRACE~\cite{li2026molecular} frames lead optimization as
resource-aware planning, CheMatAgent~\cite{wu2025chematagent} learns tool-use
policies, and Mozi~\cite{cao2026mozi} represents governed autonomy through
state-aware skill graphs.
The planner should therefore be treated as a policy over admissible scientific actions, not as a free-form text generator that happens to call tools.

The planning loop closes at different levels.
At L2, agents propose candidates or workflows, evaluate them with predictors or simulators, and refine the next action, as in dZiner~\cite{ansari2024dzinerrationalinversedesign}, ChatMOF~\cite{kang2024chatmof}, and MDCrow~\cite{campbell2026mdcrow}.
At L3, Coscientist~\cite{boiko2023autonomous} and ORGANA~\cite{darvish2025organa}
demonstrate planning and physical execution with feedback-aware recovery,
perception, or analysis. LLM-RDF~\cite{ruan2024automatic} and robotic
ChemAgents~\cite{song2025multiagent} demonstrate the stronger iterative form of
L3 because measured reaction yields or catalyst performance determine a
subsequent physical experiment. AutoLabs~\cite{panapitiya2026autolabs} remains
L2 under our evidence rule: it evaluates multi-agent protocol generation and
self-correction, but reports a critical human verification step before
instrument execution. Tippy~\cite{fehlis2025accelerating,fehlis2025technical}
also remains L2 because its reported evaluation does not establish autonomous
physical execution.
At L4, planning would become research-program design, where agents select problems, allocate resources, compare campaigns, and extract reusable rules.

\subsubsection{Memory}

Memory gives the controller persistence beyond the current prompt.
Molecular discovery may involve repeated molecule edits, failed tool calls, changing constraints, simulation parameters, synthesis attempts, assay evidence, and human decisions.
A context window can hold part of this state, but it does not provide durable, searchable, provenance-aware scientific memory.
ChemAgent~\cite{tang2025chemagent} uses self-updating memories to improve chemical reasoning, while modular drug-discovery agents~\cite{ock2026large} and TRACE~\cite{li2026molecular} show why long-horizon tasks require preserving intermediate state and action history.

Molecular memory is structured rather than conversational.
It may store molecules, conformers, pockets, score vectors, docking poses, reaction conditions, failed candidates, tool parameters, literature snippets, assay records, protocol versions, and design rationales.
ChemAgent~\cite{tang2025chemagent} is a representative typed-memory architecture with planning, execution, and knowledge memories; DrugPilot~\cite{li2025drugpilot} maintains a parameterized memory pool for multimodal drug-discovery information; and El Agente~\cite{zou2025agente} uses memory inside a hierarchical quantum-chemistry workflow.

Retrieval-augmented generation is another form of external memory.
Chemist-X~\cite{chen2023chemist} retrieves literature and database evidence for reaction-condition recommendation, ChatDrug~\cite{liu2024conversational} combines retrieval with domain feedback for conversational drug editing, and RAG-enhanced collaborative agents~\cite{lee2026rag} use retrieval for drug-discovery reasoning.
Agent-based learning from literature~\cite{ansari2024agent} extracts structured materials data from papers, while DrugAgent~\cite{inoue2025drugagent} and DrugMCTS~\cite{yang2025drugmcts} illustrate knowledge-graph and retrieval memory for drug-target or repurposing tasks.
Provenance is essential: the agent should retain a fact together with its source and any later contradictory evidence.

Experience memory connects agent design to optimization.
Previous failures can prevent repeated invalid edits, bad docking setups, weak analogs, and tool-instruction errors.
TRACE~\cite{li2026molecular} stores instruction and result histories as action-level experience, Augmented Memory~\cite{guo2024augmented} uses experience replay for sample-efficient de novo design, ExLLM~\cite{ran2025exllm} uses experience-enhanced optimization, and Mozi~\cite{cao2026mozi} stores reusable procedures in skill graphs.
Memory also introduces risks: stale literature, noisy proxy scores, contaminated examples, and incorrect tool outputs can be amplified if the controller does not decide what to write, retrieve, forget, and trust.
These memory functions support progressively longer and more reproducible
workflows, but stored context alone does not raise autonomy unless it changes a
later scientific decision.

\subsubsection{Reflection and Self-Correction}

Reflection converts feedback into correction.
In general LLM agents, reflection often means textual self-critique.
For molecular agents, this is insufficient because a fluent critique may miss an invalid structure, infeasible reaction, wrong unit, malformed docking input, unstable simulation, or unsafe protocol.
A molecular controller should therefore reflect against external signals such as validity checks, property predictors, docking, retrosynthesis, spectra, assays, failed tool calls, hardware feedback, and expert review.

A major use is iterative molecular editing and optimization.
ChatDrug~\cite{liu2024conversational} combines conversational editing with retrieval and domain feedback, AgentDrug~\cite{le2024agentdrug} uses domain feedback to steer zero-shot molecular optimization, and Probe-Before-You-Edit~\cite{yang2026probe} uses structure-based feedback before editing molecules.
DrugAssist~\cite{ye2025drugassist}, GeLLM$^3$O~\cite{dey-etal-2025-gellm3o}, and ExLLM~\cite{ran2025exllm} similarly condition later proposals on observed weaknesses of earlier candidates.
In these systems, reflection is not an explanation after generation, but part of the optimization dynamics.

Another pattern separates generation from verification.
MT-Mol~\cite{kim2025mt} uses specialized tool-based roles including verifier and reviewer functions; ChemActor~\cite{zhang2025chemactor} adds structured action sequences and multi-round review for synthesis extraction; and ChemLabs~\cite{xu2025chemlabs} uses multi-agent checking for multimodal chemistry reasoning.
Debate is a stronger variant: Mol-Debate uses disagreement among agents to improve molecular structural reasoning~\cite{zhang2026mol}, and collaborative expert LLMs expose trade-offs in multi-objective optimization~\cite{yu2025collaborative}.
However, debate is useful only when agents bring diverse evidence, tools, or objectives; otherwise it can amplify shared errors.

Reflection also applies to tool failures.
Invalid SMILES strings, malformed arguments, missing database fields, incompatible files, unstable simulations, and contradictory scorers can all corrupt downstream planning.
ChemHAS applies self-correction at the chemistry-tool interface~\cite{li2025chemhas}, and TRACE reuses previous tool-instruction failures to refine future actions~\cite{li2026molecular}.
AutoLabs~\cite{panapitiya2026autolabs} extends self-checking to the generation
of hardware-ready experimental procedures, but its reported pre-execution
human verification gate keeps the evaluated workflow at L2 under our rubric.
The autonomy contribution of reflection therefore depends on the source of the
feedback and on which later decision it is allowed to revise, not on the
presence of a self-critique step.

\subsubsection{Multi-Agent Collaboration}

Multi-agent collaboration distributes controller functions across specialized roles.
This design is natural because discovery spans heterogeneous artifacts and skills: literature, structures, protein pockets, reaction schemes, simulations, spectra, protocols, and lab readouts.
A single controller may struggle to maintain all contexts, whereas specialized agents can separate literature search, generation, scoring, synthesis planning, analysis, verification, and human communication.
The relevant design criterion is whether the division of labor matches the molecular task, not the number of agents.

Role-specialized pipelines decompose discovery into stages.
PharmAgents~\cite{gao2025pharmagents}, Prompt-to-Pill~\cite{vichentijevikj2026prompt}, MADD~\cite{solovev2025madd}, FROGENT~\cite{pan2026frogentendtoendfullprocessdrug}, and M$^4$olGen~\cite{li2026m} assign different agents or stages to target analysis, generation, scoring, synthesis, and reporting.
At L3, agents can map directly onto laboratory roles. Coscientist~\cite{boiko2023autonomous} coordinates specialized planning, search, code, and automation modules. LLM-RDF~\cite{ruan2024automatic} assigns agents to literature scouting, experiment design, hardware execution, spectrum analysis, separation instruction, and result interpretation, while robotic ChemAgents~\cite{song2025multiagent} coordinates literature, experiment design, computation, and robotic execution in an iterative materials campaign. ORGANA~\cite{darvish2025organa} combines task planning, visual feedback, robot control, and reporting. AutoLabs~\cite{panapitiya2026autolabs} and Tippy~\cite{fehlis2025accelerating,fehlis2025technical} define useful laboratory-facing roles but remain L2 here because their reported evaluations retain a pre-execution human gate or do not establish autonomous physical execution.

Many systems use an orchestrator or supervisor.
Mozi~\cite{cao2026mozi} maintains governed autonomy through state-aware skill
graphs. MASTER~\cite{rothfarb2025hierarchical} uses hierarchical multi-agent
reasoning for functional-materials discovery. El Agente~\cite{zou2025agente}
organizes quantum-chemistry workflows through hierarchical control.
This authority structure improves traceability, since decisions can be attributed to a supervisor, specialist, tool, or human gate, and it supports permissioning for high-risk actions.

Collaboration can also improve critique~\cite{zheng2024picture}.
Mol-Debate~\cite{zhang2026mol}, MT-Mol~\cite{kim2025mt}, ChemLabs~\cite{xu2025chemlabs}, and MASTER~\cite{rothfarb2025hierarchical} use multi-agent disagreement, verification, or peer review as quality control.
Agents-on-a-Tree~\cite{zhang2026agents}, DrugMCTS~\cite{yang2025drugmcts}, multi-GPT-agent reinforcement learning~\cite{hu2023novo}, and collaborative expert LLMs~\cite{yu2025collaborative} coordinate agents over shared chemical search spaces.
Humans can also be first-class collaborators: collaborative structure-based drug design~\cite{gao2025pushing} and ORGANA~\cite{darvish2025organa} keep experts in the loop for objectives, approvals, preferences, and final judgment.
Role specialization can support any level; the classification depends on the
strongest evidence-conditioned action performed by the team as a whole.

\subsubsection{Cross-Cutting Synthesis}

Figure~\ref{fig:agent-framework-heatmap} contains 96 faculty-level assignments: 39 at L1, 48 at L2, and 9 at L3. Fixed tool chains, predefined role pipelines, and review-only ensembles remain L1 unless evidence changes a later scientific decision. Feedback-aware physical workflows qualify as L3, with measurement-driven experiment selection identified as iterative L3. No surveyed paper meets the L4 criterion.

The five faculties are therefore enabling mechanisms rather than autonomy levels. Their role changes with feedback source and action authority: memory progresses from retrieval to experimental state, while collaboration progresses from review ensembles to divisions of laboratory work. As authority increases, permissioning, traceability, tool constraints, and approval gates become part of the controller itself.

%% file: sections/03framework-03-Tool.tex
\subsection{Domain-Specific Molecular Toolboxes}

Molecular toolboxes translate language-level plans into executable chemical operations over structures, databases, simulations, spectra, and laboratory protocols.
ChemCrow~\cite{m2024augmenting} and CACTUS~\cite{mcnaughton2024cactus} illustrate chemistry-tool control, Coscientist~\cite{boiko2023autonomous} extends it toward experiments, and ToolUniverse~\cite{gao2025tooluniverse} treats tool composition as reusable scientific infrastructure. 
The toolbox is therefore part of the agent architecture: it defines both the available actions and the observations that can enter the control loop.

Those observations differ in cost and reliability. Computational scores, assay records, spectra, and robot logs must be interpreted within their operating conditions. Tool access alone is not beneficial~\cite{yu2025tooling}; self-correction, experience, and workflow structure determine whether evidence improves a later decision~\cite{li2025chemhas,li2026molecular,molclaw2026}.
Accordingly, fixed calls remain L1, evidence-adaptive computational tool use is L2, and autonomous physical execution with feedback is L3. The criterion is scientific authority exercised through tools, not the number of tools.

\subsubsection{Molecular state construction and knowledge grounding}

\paragraph{Entity resolution and molecular state construction.}

A molecular agent first has to determine which chemical object it is acting on. User inputs may name a compound, give an IUPAC name, provide a CAS number, mention a target or protein family, quote a paper fragment, or describe an assay. The agent must resolve these surface forms into identifiers and representations such as SMILES, InChI, PubChem CID, ChEMBL ID, UniProt ID, PDB structures, or assay records. OPSIN, PubChem, ChEMBL, RCSB PDB, UniProt, ZINC, and Materials Project belong in this grounding layer~\cite{lowe2011opsin,kim2023pubchem,zdrazil2024chembl,berman2000pdb,uniprot2023uniprot,irwin2020zinc20,jain2013commentary}. A wrong molecule, target, protonation state, or assay condition can invalidate the rest of the workflow. ChemCrow~\cite{m2024augmenting} and CACTUS~\cite{mcnaughton2024cactus} rely on name-to-structure conversion and database grounding as entry points for tool use. ChemAgent~\cite{tang2025chemagent}, Chemist-X~\cite{chen2023chemist}, and LLaMP~\cite{chiang2025llamp} show how structured state construction can be combined with memory or retrieval.

\paragraph{Cheminformatics validation and descriptor computation.}

Cheminformatics toolkits provide many of the low-level operations that molecular agents need. RDKit~\cite{landrum2016rdkit} and Open Babel~\cite{oboyle2011openbabel} parse SMILES and other formats, canonicalize structures, validate valence and aromaticity, preserve or check stereochemistry, compute descriptors, generate fingerprints, support similarity search, and convert file formats. These operations matter because LLM-generated molecular strings can be fragile, invalid, or underspecified. Once a text output becomes a checkable molecular object, the tool can act both as calculator and validator. ChemCrow~\cite{m2024augmenting} and CACTUS~\cite{mcnaughton2024cactus} use such operations for chemistry problem solving. MT-Mol~\cite{kim2025mt}, ToolMol~\cite{zhou2026toolmol}, CheMatAgent~\cite{wu2025chematagent}, TRACE~\cite{li2026molecular}, and MolClaw~\cite{molclaw2026} place them inside longer optimization or tool-planning loops. Evaluation can draw on established benchmarks for molecular design and property prediction, including MoleculeNet~\cite{wu2018moleculenet}, GuacaMol~\cite{brown2019guacamol}, MOSES~\cite{polykovskiy2020moses}, and TDC~\cite{huang2021tdc}.

\paragraph{Database and literature grounding.}

Molecular agents also need structured scientific memory. Databases and literature tools provide evidence about identity, target annotations, protein structures, bioactivity values, assay metadata, commercial availability, reaction precedent, and experimental results. Molecular databases differ from general web search because they preserve identifiers, units, curation history, and provenance. PubChem and ChEMBL ground compounds and bioactivity data; RCSB PDB and UniProt connect agents to protein structures and protein knowledge; ZINC and Materials Project support purchasable-molecule and materials retrieval~\cite{kim2023pubchem,zdrazil2024chembl,berman2000pdb,uniprot2023uniprot,irwin2020zinc20,tingle2023zinc,jain2013commentary}. Chemist-X~\cite{chen2023chemist} uses retrieved evidence for reaction-condition recommendation. LLaMP~\cite{chiang2025llamp} grounds materials reasoning in high-fidelity retrieval, and agent-based literature learning extracts structured materials datasets from papers~\cite{ansari2024agent}. Retrieval-augmented drug-discovery agents extend this pattern to drug-target reasoning and collaborative molecular decisions~\cite{inoue2025drugagent,lee2026rag}. The difficult parts are entity linking, unit normalization, conflicting evidence, and versioned provenance.

\subsubsection{In-silico evaluation and simulation tools}

\paragraph{Low-cost oracles and structure-based screening.}

In-silico evaluation tools let an agent test a hypothesis before synthesis or wet-lab work. Low-cost tools include descriptor calculators, drug-likeness filters, synthetic-accessibility scores, ADMET predictors, and toxicity predictors. Structure-based tools add docking, scoring, binding-pose analysis, and virtual screening. AutoDock Vina, GNINA, and DiffDock, together with benchmarks such as PDBbind and CASF, are common substrates for pose prediction, scoring, ranking, and screening~\cite{trott2010autodock,mcnutt2021gnina,corso2023diffdock,wang2005pdbbind,su2018comparative}. These tools make iterative optimization practical: the agent proposes candidates, evaluates them with proxy objectives, and revises them. CACTUS~\cite{mcnaughton2024cactus} and MT-Mol~\cite{kim2025mt} incorporate tool-based evaluation into agent workflows. Several drug-discovery agents combine generation, filtering, docking, retrieval, and optimization in more specialized settings~\cite{zhou2026toolmol,le2024agentdrug,yang2026probe,solovev2025madd,li2025drugpilot,yang2025drugmcts,gao2025pushing,molclaw2026}. These scores should still be treated as decision signals, not as ground truth. Docking scores approximate binding, ADMET models depend on training distributions, and heuristic filters can reject unusual but useful chemistry.

\paragraph{Quantum chemistry and atomistic simulation.}

Quantum-chemical and atomistic simulation tools provide physical feedback at higher cost. They compute geometries, single-point energies, vibrational frequencies, reaction energetics, electronic structures, and material properties. ASE, xTB, Psi4, ORCA, Gaussian, pymatgen, and LAMMPS expose these calculations through programmable interfaces, which makes them usable inside agent workflows~\cite{larsen2017ase,bannwarth2019gfn2,bannwarth2021xtb,parrish2017psi4,neese2020orca,frisch2016gaussian16,ong2013pymatgen,thompson2022lammps}. For molecular agents, these tools ground part of the reasoning in physics rather than text or heuristic scores. El Agente~\cite{zou2025agente} demonstrates autonomous quantum chemistry. ChemReasoner~\cite{sprueill2024chemreasoner} uses quantum-chemical feedback to guide heuristic search over an LLM's chemistry knowledge space. ChemGraph exposes molecular simulation workflows as agent-compatible computational chemistry tasks~\cite{pham2026chemgraph}. Monte Carlo Thought Search~\cite{sprueill2023monte} explores reasoning paths in catalyst design, and CheMatAgent~\cite{wu2025chematagent} learns tool-use policies for chemistry and materials science. These systems reveal a demanding control problem: the agent must choose method, basis set, charge, spin state, solvent model, convergence criterion, and computational budget. Materials-agent surveys make a similar argument for atomistic and materials workflows~\cite{zhang2026agenticmaterials}.

\paragraph{Molecular dynamics and free-energy workflows.}

Molecular dynamics tools move agents from static structures to time-dependent behavior. They support studies of protein flexibility, ligand stability, solvent effects, conformational transitions, and binding-related dynamics. GROMACS, OpenMM, AmberTools, and LAMMPS provide the computational substrate for many of these workflows~\cite{abraham2015gromacs,eastman2017openmm,case2023ambertools,thompson2022lammps}. Free-energy workflows add methods such as FEP, TI, MM/PBSA, ABFE, umbrella sampling, and related protocols for ranking candidates. MDCrow~\cite{campbell2026mdcrow} automates molecular-dynamics workflows, including setup, execution, and analysis. DynaMate~\cite{guilbert2025dynamate} targets autonomous protein and protein-ligand MD workflows, while MDAgent2~\cite{shi2026mdagent2} studies code generation, execution, evaluation, and self-correction for MD simulations. ToolMol~\cite{zhou2026toolmol}, MADD~\cite{solovev2025madd}, TRACE~\cite{li2026molecular}, modular task-execution agents~\cite{ock2026large}, and MolClaw~\cite{molclaw2026} combine MD or free-energy-like validation with docking, ADMET, and optimization. Evaluation has to consider stability, convergence, sampling sufficiency, cost, and whether simulation feedback changes downstream decisions.

\subsubsection{Synthesis, characterization, and experimental action tools}

\paragraph{Reaction prediction and synthesis feasibility.}

A molecular candidate is not actionable until the agent can connect it to feasible synthesis, available starting materials, and executable reaction conditions. Reaction-prediction and retrosynthesis tools estimate whether a proposed molecule can be made, how many steps may be required, which reagents or catalysts are plausible, and whether route constraints match the intended application. AiZynthFinder and ASKCOS are useful reference points for this layer~\cite{genheden2020aizynthfinder,tu2025askcos}. Agentic systems increasingly treat synthesis planning as part of molecular design rather than as post-processing. ChemCrow~\cite{m2024augmenting} uses synthesis-oriented tools in chemistry problem solving. Chemist-X~\cite{chen2023chemist} focuses on reaction-condition recommendation, RETRO-R1~\cite{liu2026retro} studies agentic retrosynthesis, and Llamole~\cite{liu2025multimodal} integrates inverse molecular design with retrosynthetic planning. ChemActor~\cite{zhang2025chemactor} converts synthesis procedures into structured chemical action sequences. Together, these works move the action space from attractive structures to candidates that can plausibly enter a Design, Make, Test, Analyze cycle.

\paragraph{Characterization, spectroscopy, and experimental readout.}

Closed-loop molecular agents must read experimental evidence, not just propose experiments. Characterization and spectroscopy tools convert measurements into structured feedback: NMR assignments, MS or LC-MS peaks, IR or UV spectra, XRD patterns, XANES features, microscopy images, plots, and assay readouts. The NIST Chemistry WebBook, MassBank, NMRShiftDB, and FDMNES provide representative infrastructure for thermochemical, spectral, NMR, MS, and XANES analysis~\cite{nist2025chemwebbook,horai2010massbank,neumann2026massbank,steinbeck2003nmrshiftdb,kuhn2024nmrshiftdb2,joly2001xanes,guda2015optimizedfdmnes}. This feedback tells the agent whether a reaction succeeded, whether the expected product formed, whether impurities appeared, and whether the next experiment should continue, change, or stop. LLM-RDF~\cite{ruan2024automatic}, ORGANA~\cite{darvish2025organa}, and robotic ChemAgents~\cite{song2025multiagent} demonstrate experimental execution and readout interpretation. Tippy~\cite{fehlis2025accelerating,fehlis2025technical} specifies corresponding laboratory-facing roles, but its reported evidence does not establish autonomous physical execution. ChemGraph-XANES~\cite{grizzi2026chemgraphxanes} gives a concrete XANES simulation and analysis workflow for agents, while ChemLabs~\cite{xu2025chemlabs} examines multimodal reasoning in chemistry. The central issue is uncertainty: spectra and curves can support several hypotheses, so the agent should preserve alternatives instead of selecting one explanation too early.

\paragraph{Laboratory automation and safety gates.}

Laboratory automation tools move molecular agents from recommendation to physical action. They include protocol generators, robotic synthesis platforms, liquid-handling systems, cloud laboratories, reaction-execution interfaces, and instrument-control tools. This step creates stronger governance requirements. A failed computational job wastes time; a failed laboratory action can waste material, damage instruments, create unsafe conditions, or violate compliance rules. Coscientist~\cite{boiko2023autonomous}, LLM-RDF~\cite{ruan2024automatic}, ORGANA~\cite{darvish2025organa}, and robotic ChemAgents~\cite{song2025multiagent} combine execution, characterization, and feedback in physical workflows. AutoLabs~\cite{panapitiya2026autolabs} instead evaluates hardware-ready protocol generation with human verification before execution, while Tippy~\cite{fehlis2025accelerating,fehlis2025technical} contributes a laboratory-facing architecture. Safety gates can include controlled-chemical checks, protocol validation, resource constraints, and human approval before high-risk actions, as suggested by safety-aware tool use in ChemCrow~\cite{m2024augmenting}.

\subsubsection{Tool orchestration and evaluation}

\paragraph{Workflow orchestration, state passing, and provenance.}

Molecular agency becomes most visible when tools are linked into workflows. An agent may pass a molecule from SMILES to SDF, convert a structure into docking input, prepare a PDB file for MD, summarize a trajectory into stability metrics, match spectra against candidate structures, or translate a reaction plan into a robot-executable protocol. Each transition can lose stereochemistry, protonation state, units, conformers, file-format details, or experimental conditions. ChemCrow~\cite{m2024augmenting} and CACTUS~\cite{mcnaughton2024cactus} demonstrate early chemistry tool orchestration. ChemHTS~\cite{li2025chemhts}, ChemHAS~\cite{li2025chemhas}, CheMatAgent~\cite{wu2025chematagent}, Mozi~\cite{cao2026mozi}, and MolClaw~\cite{molclaw2026} study more structured forms of tool stacking, tool learning, hierarchical skills, or governed autonomy. Longer-horizon systems such as El Agente~\cite{zou2025agente}, ChemGraph~\cite{pham2026chemgraph}, ChemGraph-XANES~\cite{grizzi2026chemgraphxanes}, MDCrow~\cite{campbell2026mdcrow}, DynaMate~\cite{guilbert2025dynamate}, MDAgent2~\cite{shi2026mdagent2}, and LLaMP~\cite{chiang2025llamp} point to the need for state tracking, logging, versioning, intermediate-memory management, and provenance records. ToolUniverse points in the same direction for general scientific tooling, where standardized specifications and reusable composition become infrastructure~\cite{gao2025tooluniverse}.

\paragraph{Failure recovery, cost control, and calibration.}

Tool use also introduces failure modes that are specific to chemistry. A workflow can fail because of invalid SMILES, lost stereochemistry, missing hydrogens, inconsistent protonation states, malformed docking inputs, DFT non-convergence, unstable MD trajectories, database mismatches, unit errors, unavailable reagents, or unsafe protocols. A robust molecular agent should detect these failures, explain them, retry with corrected inputs, switch tools when appropriate, or request human intervention. Tooling-or-Not-Tooling~\cite{yu2025tooling} motivates this caution by showing that tools can be harmful when used in the wrong setting. ChemHAS~\cite{li2025chemhas}, TRACE~\cite{li2026molecular}, and MolClaw~\cite{molclaw2026} suggest that self-correction, action-level experience, and workflow-level skills can improve tool-facing behavior. Computational workflow agents such as El Agente~\cite{zou2025agente}, ChemGraph~\cite{pham2026chemgraph}, MDCrow~\cite{campbell2026mdcrow}, DynaMate~\cite{guilbert2025dynamate}, and MDAgent2~\cite{shi2026mdagent2} make cost control especially important, since high-cost DFT, MD, or free-energy calculations should not run when a lower-fidelity check is enough. Calibration matters for the same reason. The agent should know when a proxy score is useful for ranking, when higher-fidelity validation is needed, and when the evidence remains inconclusive.

\paragraph{Evaluation across the autonomy ladder.}

Toolbox evaluation should retain a tool-specific view even when the detailed benchmark taxonomy is consolidated in Section~\ref{sec:evaluation_benchmarking}. Beyond final-answer accuracy, it should test molecular identity preservation, valid calls, state passing, recovery, provenance, and whether tool evidence improves a later decision. MolViBench~\cite{li2026molvibench}, MolBench~\cite{molclaw2026}, and MatTools~\cite{liu2025mattools} expose parts of this process, while L3 evaluation additionally requires protocol executability, hardware compatibility, safety, and faithful interpretation of measurements.

%% file: sections/03framework-04-Optimization.tex
\subsection{Learning and Optimization}
\label{sec:learning-optimization}

Learning and optimization describe two complementary dimensions of improvement in molecular agents.
\textbf{Learning concerns what the agent acquires or updates from experience}, including policies, memories, predictive models, acquisition strategies, tool-use behaviors, and reusable workflows.
\textbf{Optimization concerns what component or decision variable is deliberately improved with respect to an objective}, ranging from molecular candidates and experimental conditions to action sequences, tool choices, search policies, and the workflow itself.

The distinction is therefore not simply between improvement within a task and improvement across trajectories.
Optimization may operate within a single trajectory or across many episodes, while learning may occur online during a trajectory or accumulate across repeated tasks and campaigns.
The two processes also interact: optimization can exploit previously learned knowledge, while learning itself is often driven by an optimization objective.

For molecular agents, the important question is not whether a system contains a learning algorithm or repeatedly performs optimization.
Rather, the evidence for agency lies in whether observations, evaluations, or accumulated experience change a later consequential scientific decision.
The autonomy level then depends on \emph{what kind of decision is changed}: a fixed computational procedure remains L1; computational evidence that changes a later candidate, plan, tool call, or stopping decision supports L2; physical execution with incorporated experimental feedback supports L3; and accumulated evidence that changes objectives, strategy, or scientific agenda across campaigns is required for L4.

\begin{figure*}[t]
\centering
\includegraphics[width=1.0\textwidth]{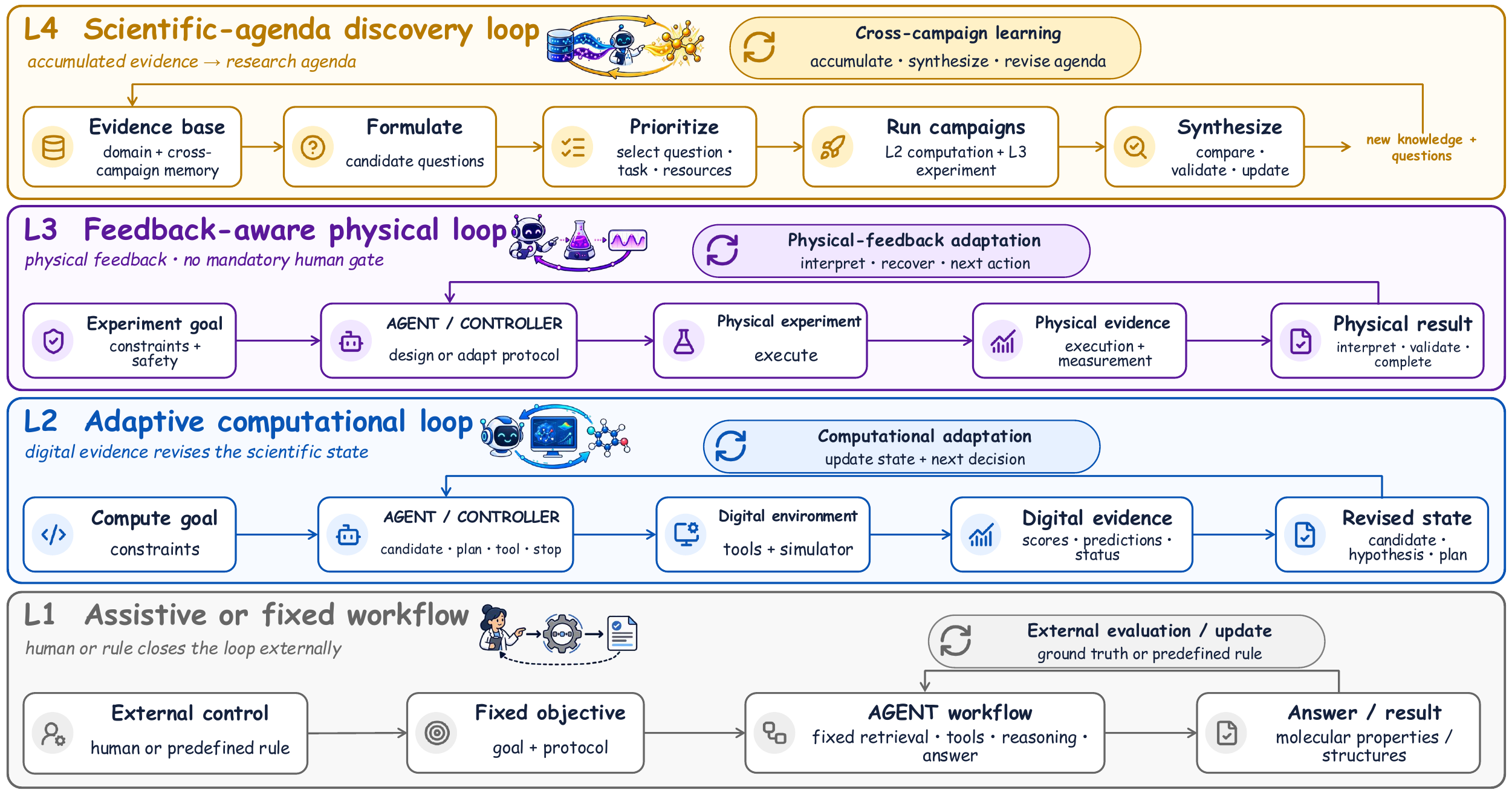}
\caption{Hierarchical learning and optimization in molecular LLM agents.
Optimization may target candidates, actions, workflows, or experiments, while learning updates policies, memories, models, strategies, and reusable workflows from experience.
These mechanisms can operate at multiple autonomy levels: from fixed computational procedures (L1), through feedback-driven computational decisions (L2) and physical experimentation (L3), to cross-campaign learning that revises higher-level scientific objectives or agendas (L4).}
\Description{Learning and optimization operate across multiple levels of molecular-agent autonomy, from fixed computational procedures to feedback-driven computational decisions, physical experimentation, and cross-campaign learning.}
\label{fig:learning-optimization-framework}
\end{figure*}

\subsubsection{Optimization Targets and Search Spaces}

Early molecular learning and optimization systems focused primarily on candidate-level decisions, such as generating a molecule or editing a molecular graph.
More recent molecular agents broaden the object of optimization from the molecular candidate to the discovery process itself.
They may optimize which candidate to evaluate, which tool to invoke, which evidence to trust, how computational resources are allocated, when a failed action should be repaired, and whether a multi-step trajectory should continue or stop.

Section~\ref{sec:molecular-perception} describes molecular representations in detail; here, the relevant issue is the candidate and action spaces they expose to an optimizer.
Sequence policies generate or edit strings, whereas graph policies act on atoms, bonds, fragments, or scaffolds.
Property-directed SMILES generation in Molecular De Novo Design~\cite{olivecrona2017molecular} and ReLeaSE~\cite{popova2018deep} established the propose--score--update pattern.
REINVENT4 extends this paradigm through transfer learning, reinforcement learning, curriculum learning, and multi-component scoring~\cite{loeffler2024reinvent4}.

String actions integrate readily with language models but can fail at ring closures, stereochemistry, or syntax.
SELFIES, Group SELFIES, and SAFE reduce some syntactic failures through constraints or chemically meaningful units~\cite{krenn2020selfies,cheng2023group,noutahi2023safe}, although validity does not imply stability or synthesizability.
Graph actions instead make molecular topology and local edits explicit.
GCPN constructs graphs under validity constraints, whereas MolDQN performs local lead-optimization edits from an existing molecule~\cite{you2018graph,zhou2019optimization}.

Representation determines what can be changed, but it does not determine whether the system is agentic.
The same optimization policy may appear inside a fixed human-controlled pipeline or an adaptive agent.
What matters is whether feedback from an earlier action alters a later proposal, evaluation strategy, tool choice, or selection decision.

\subsubsection{Multi-Objective Decision Making}

Evaluation provides the objectives and constraints under which optimization proceeds.
Early molecular optimization studies often emphasized a single objective, such as QED, LogP, molecular similarity, predicted binding affinity, or docking score.
Drug discovery and materials design, however, generally require simultaneous consideration of activity, selectivity, toxicity, ADMET, stability, synthetic accessibility, novelty, diversity, and cost.

Because these criteria may conflict, optimization commonly relies on composite objectives, constrained optimization, threshold-based filtering, Pareto selection, or combinations of these mechanisms.
DrugEx v2~\cite{liu2021drugex} combines reinforcement learning with Pareto ranking to balance multiple molecular properties rather than maximize a single weighted score.
QADD~\cite{qadd2023} performs iterative multi-objective reinforcement learning for de novo drug design, while MARS~\cite{xie2021mars} combines Markov chain Monte Carlo with graph editing in a propose--evaluate--retain loop.

Multi-objective decision making is more than assigning a score.
It determines which trade-offs are acceptable, which constraints are non-negotiable, and which candidates receive additional computational or experimental resources.
Some objectives can be balanced against one another, whereas synthesis feasibility, safety, or executability may instead be enforced as hard constraints.

These mechanisms remain optimization methods rather than evidence of autonomy by themselves.
They contribute to an L2 agent only when their evaluations affect a subsequent candidate, edit, tool invocation, resource-allocation decision, or stopping condition without an intervening human decision.
Likewise, applying an optimizer to an experimental objective does not establish L3 unless physical execution and incorporated experimental feedback are demonstrated.

Most computational optimization relies on surrogate objectives, including property predictors, docking scores, estimated binding affinity, synthetic accessibility, ADMET models, and other low-cost approximations.
These signals make repeated evaluation feasible, but they are not equivalent to the scientific outcomes ultimately sought.

A sufficiently capable optimizer may therefore exploit weaknesses in a proxy rather than discover a scientifically useful candidate.
Such failure becomes especially consequential in long-running agentic loops because biased feedback can affect many subsequent decisions.
Robust optimization should consequently incorporate validity checks, uncertainty estimates, applicability-domain analysis, diversity constraints, and higher-fidelity confirmation where appropriate, while explicitly distinguishing predicted evidence from measured evidence.

\subsubsection{Exploration and Acquisition}

Optimization also requires deciding where additional search or evidence is most valuable.
Chemical space is extremely large, and useful regions are sparse.
An optimizer that exploits current rewards too aggressively may converge around a narrow family of high-scoring structures and repeatedly make minor local modifications, reducing scaffold diversity and increasing vulnerability to biased surrogate objectives.

Mol-AIR~\cite{molair2024} introduces adaptive intrinsic rewards for goal-directed molecular generation, encouraging novelty and broader coverage in addition to target-property performance.
Augmented Hill-Climb~\cite{thomas2022augmented} improves the sample efficiency of REINVENT-like molecular language models, reducing the number of unproductive candidates requiring downstream evaluation.

When evaluations are expensive, acquisition itself becomes an optimization problem.
Phoenics~\cite{hase2018phoenics} uses previous observations and predictive uncertainty to recommend candidates or conditions expected to provide either high objective value or useful information.
The constrained latent-space approach of G\'omez-Bombarelli et al.~\cite{gomez2018automatic} similarly provides a smoother domain in which to conduct search than direct optimization over discrete strings or graphs.

Exploration becomes agentic when accumulated evidence changes where the controller searches, what information it chooses to acquire, how much resource it allocates to an evaluation, or when it decides that further search is no longer worthwhile.
This distinction becomes increasingly important as the workflow moves from inexpensive computational scoring toward simulations and physical experiments.

\subsubsection{Agent Learning and Workflow Optimization}

Molecular agents extend improvement beyond the candidate itself.
An agent may \emph{learn} which tools are reliable, which representations are useful, which failures recur, which search regions are productive, or which workflows succeed under particular objectives.
It may then \emph{optimize} later behavior using this acquired information by changing its policy, tool sequence, resource allocation, repair strategy, or stopping rule.

ReMol~\cite{wang2025remol} combines LLM guidance with reinforcement learning.
The language model contributes chemical priors and reasoning signals, while reinforcement learning updates the molecular policy using property feedback.
This illustrates how learned knowledge can influence optimization without restricting the LLM to direct molecular-string generation.

ChemCRAFT~\cite{li2026agentic} learns tool-use policies from trajectories in chemical sandboxes.
MolClaw~\cite{molclaw2026} organizes tool-level, workflow-level, and discipline-level skills for molecular evaluation, screening, and optimization.
These systems illustrate learning at the level of agent behavior rather than only at the level of molecular generation.

General agent-learning methods provide related mechanisms.
Reflexion~\cite{shinn2023reflexion} stores natural-language feedback that can alter behavior in later attempts without parameter updates.
Agent Lightning~\cite{agentlightning2025} represents multi-step agent execution as a Markov decision process and applies reinforcement learning to assign credit across trajectories.

The existence of memory, reinforcement learning, or trajectory storage nevertheless does not establish scientific agency.
Learning is scientifically consequential only when acquired information is reused to change a later decision, and when the effect is demonstrated beyond the episode from which the information was obtained.
Likewise, workflow optimization requires evidence that feedback changes a consequential candidate, hypothesis, plan, tool choice, resource-allocation decision, or stopping rule.
Merely executing a predefined sequence of tools, storing a transcript, or correcting tool syntax remains compatible with L1.

\input{tables/benchmark}

\subsubsection{Laboratory-in-the-Loop Optimization}

Laboratory-in-the-loop optimization replaces or supplements computational proxies with physical observations.
The optimization target may be a molecular candidate, reaction condition, formulation, synthesis protocol, or experimental sequence, while learning can update the acquisition policy, predictive model, experimental strategy, or workflow from observed outcomes.

Experimental feedback is typically slower, noisier, and more expensive than computational scoring.
The agent must therefore balance target performance with information gain, material use, executability, reproducibility, uncertainty, and safety.

Bayesian reaction optimization~\cite{shields2021bayesian} demonstrates how previous observations and uncertainty can guide the selection of subsequent experimental conditions.
The method provides an experimental learning and optimization mechanism, although that mechanism alone does not establish an LLM-centered autonomous agent.

Self-driving laboratories provide clearer evidence of closed physical loops.
AlphaFlow~\cite{alphaflow2023} uses reinforcement learning to guide a microfluidic platform in the exploration and optimization of multi-step chemical processes.
Autonomous Polymer Synthesis~\cite{autopolymer2022} performs multi-objective closed-loop optimization for polymer synthesis under Pareto trade-offs.
In both cases, measured outcomes affect subsequent experimental decisions.

LLM-centered systems increasingly connect planning with laboratory execution.
Coscientist~\cite{boiko2023autonomous} plans and executes experiments through equipment or cloud-laboratory interfaces and uses feedback to interpret measurements or repair execution code.
ORGANA~\cite{darvish2025organa} combines agent-generated plans with visual feedback during physical execution.
AutoLabs~\cite{panapitiya2026autolabs} evaluates self-corrected, hardware-ready protocol generation, but a mandatory human verification step before execution prevents the reported workflow from satisfying L3 under our criterion.

A stronger form of L3 occurs when a measurement determines another experiment.
LLM-RDF~\cite{ruan2024automatic} uses measured reaction yields to select subsequent reaction conditions.
Robotic ChemAgents~\cite{song2025multiagent} uses measured catalyst performance to select later compositions for physical validation.
The Mobile Robotic Chemist~\cite{burger2020mobile} and A-Lab~\cite{szymanski2023lab} likewise use observed experimental outcomes to choose subsequent physical actions.

These cases distinguish broad L3 from iterative L3.
Broad L3 requires autonomous physical execution together with incorporated execution or measurement feedback that changes recovery, interpretation, or completion decisions.
Iterative L3 requires the stronger condition that a measured result determines a subsequent physical experiment or experimental condition.

Human monitoring and emergency stopping do not necessarily reduce the autonomy level, whereas a mandatory approval gate before routine physical execution does.
Fixed protocol replay remains L1, and a proposed laboratory architecture without demonstrated physical execution does not establish L3.

\subsubsection{Cross-Trajectory and Cross-Campaign Learning}

Learning can persist beyond a single trajectory.
Across repeated tasks, an agent may accumulate reusable memories, update policies, estimate tool reliability, refine acquisition strategies, or induce workflow-level skills from previous successes and failures.
Such cross-trajectory learning can improve later L2 or L3 decisions without necessarily changing the scientific objective itself.

Cross-campaign learning is stronger.
Here, accumulated evidence from completed optimization or experimental campaigns alters higher-level scientific choices, such as which objective to pursue, which hypothesis to investigate, which region of chemical space deserves further study, which experimental strategy should be abandoned, or how future campaigns should be organized.

This distinction is important for L4.
Reusing a successful workflow, updating a policy from previous trajectories, or fine-tuning an agent from accumulated experience may constitute learning, but it does not by itself establish L4 autonomy.
L4 requires evidence that learning changes a consequential scientific decision at the campaign or agenda level rather than merely improving execution of a previously specified objective.

\subsubsection{What Learning and Optimization Evidence Establishes}

Learning and optimization mechanisms should therefore not be mapped directly onto autonomy levels.
Reinforcement learning, Bayesian optimization, memory, multi-objective selection, and workflow adaptation can all occur at different levels depending on how their outputs affect subsequent decisions.

A fixed generator, scorer, optimizer, or workflow remains L1 when iteration is prescribed by a human or fixed procedure.
L2 requires computational evidence or learned experience to change a later candidate, hypothesis, plan, tool call, acquisition decision, or stopping condition.
L3 additionally requires physical execution together with incorporated execution or measurement feedback; measurement-selected follow-up experiments provide the stronger iterative form.
L4 requires learning across campaigns to influence higher-level objective selection, strategy, hypothesis formation, or scientific agenda.

Current evidence therefore spans candidate and workflow optimization at L1 and L2, together with bounded physical optimization and learning at L3.
Coscientist and ORGANA~\cite{boiko2023autonomous,darvish2025organa} demonstrate feedback-aware physical workflows, while AlphaFlow, autonomous polymer synthesis, LLM-RDF, Robotic ChemAgents, the Mobile Robotic Chemist, and A-Lab~\cite{alphaflow2023,autopolymer2022,ruan2024automatic,song2025multiagent,burger2020mobile,szymanski2023lab} demonstrate measurement-driven experimental iteration.
These results show increasingly capable learning and optimization loops, but none by itself establishes the cross-campaign objective selection and scientific-agenda revision required for L4.

%% file: tables/benchmark.tex
\begin{table*}[t]
\centering
\caption{Landscape of representative molecular and scientific-agent benchmarks. The table emphasizes evaluated interaction and evidence rather than ranking benchmark-specific scores, which are not directly comparable.}
\label{tab:agent_benchmark_landscape}
\scriptsize
\setlength{\tabcolsep}{3.0pt}
\renewcommand{\arraystretch}{1.10}
\begin{tabularx}{\textwidth}{@{}>{\raggedright\arraybackslash}p{2.35cm}>{\raggedright\arraybackslash}p{4.05cm}>{\raggedright\arraybackslash}X>{\raggedright\arraybackslash}p{4.35cm}@{}}
\toprule
\textbf{Benchmark} & \textbf{Scientific scope} & \textbf{Agent interaction or loop} & \textbf{Primary evaluation signal} \\
\midrule
\textbf{MolViBench}~\cite{li2026molvibench} & 358 molecular tasks, 12 workflows, and five difficulty levels & Generates executable programs for multi-step molecular workflows & Program correctness, execution success, and degradation with workflow complexity \\
\textbf{MolBench}~\cite{molclaw2026} & Molecular screening, optimization, and end-to-end discovery challenges spanning 8--50+ tool calls & Connects filtering, affinity estimation, molecular editing, and workflow execution & Subtask quality and end-to-end challenge completion \\
\textbf{ChemCost}~\cite{wu2026chemcost} & 1,427 reactions, 2,261 chemicals, and 230,775 price quotes & Reasons over reaction components and noisy procurement information & Cost-error tolerance, robustness to noise, and component-level attribution \\
\textbf{MDGym}~\cite{kumar2026mdgym} & 169 molecular simulations and 303 tasks across two MD engines & Configures, runs, diagnoses, and repairs simulation workflows & Executable task success by difficulty, engine, and failure type \\
\textbf{ChemReason-Bench}~\cite{zhang2026chemreasonbench} & 7,306 tasks instantiated from 500 organic reactions in six formats & Produces or validates ordered, condition-aware, schema-constrained procedure steps & Ordering, constraint validation, entity-role grounding, and parseable completion \\
\textbf{Corral}~\cite{alampara2025corral} & Four chemistry and materials environments: MD, ML, catalysis, and spectroscopy & Compares tool-calling and ReAct-style agents in executable expert-designed tasks & Task success, tool-use failures, and sensitivity to task--tool alignment \\
\textbf{MADE}~\cite{malik2026made} & Closed-loop computational materials discovery over chemical systems & Proposes and evaluates candidates under a constrained oracle budget, then adapts the search & Discovery efficacy, efficiency, and scaling with search-space complexity \\
\textbf{ScienceAgent\-Bench}~\cite{chen2025scienceagentbench} & 102 tasks derived from 44 papers across four disciplines & Produces executable research code with paper and data context & Full-task and partial success, expert-knowledge dependence, and execution correctness \\
\textbf{SciAgentBench}~\cite{shen2026sciagentgym} & 259 tasks, 1,134 subquestions, and 1,780 scientific tools & Selects and composes tools from elementary calls to long workflows & Step and overall success, tool routing, and performance versus interaction horizon \\
\textbf{SciCode}~\cite{tian2024scicode} & 80 scientific coding problems and 338 expert-designed subproblems & Implements research algorithms from specifications and intermediate requirements & Subproblem and full-problem execution success \\
\textbf{CORE-Bench}~\cite{siegel2024corebench} & 270 reproducibility tasks from 90 papers in three disciplines & Reproduces published results from code, data, text, and visual artifacts & Reproduction accuracy across difficulty levels and modalities \\
\textbf{SciAgentArena}~\cite{liu2026sciagentarena} & Approximately 200 real-world scientific tasks across multiple domains & Solves interactive research scenarios with stepwise verification & Stepwise task completion and behavior in specified versus open-ended scenarios \\
\bottomrule
\end{tabularx}
\end{table*}

%% file: sections/03framework-05-Benchmarking.tex
\section{Evaluation and Benchmarking}
\label{sec:evaluation_benchmarking}

Evaluation must distinguish component competence from scientific loop closure.
Executable-workflow benchmarks can reveal whether an agent constructs valid programs, selects tools, and completes multi-step interactions correctly~\cite{li2026molvibench,shen2026sciagentgym}. Closed-loop discovery environments ask a different question: whether observations redirect a budgeted search toward better candidates~\cite{malik2026made}. A chemistry question-answering score therefore cannot establish reliable tool use, while a single end-to-end success rate can hide whether the decisive contribution came from the controller, foundation model, tool, or evaluator. We organize representative benchmarks by the interaction they expose and the scientific state they require the agent to change.

\begin{figure*}[t]
  \centering
  \includegraphics[width=\textwidth]{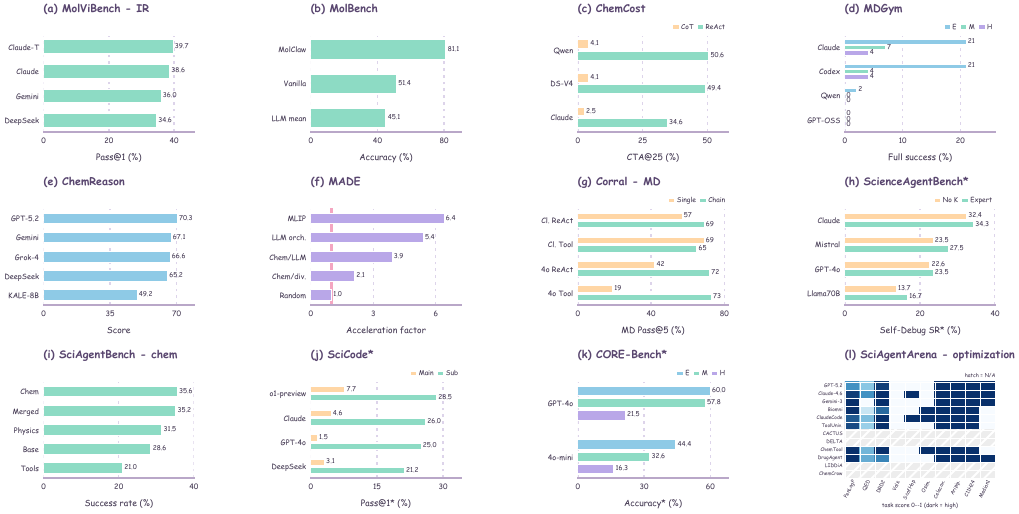}
  \caption{Compact reported result slices for all 12 benchmarks in
  Table~\ref{tab:agent_benchmark_landscape}: (a) MolViBench IR
  ~\cite{li2026molvibench}; (b) MolBench~\cite{molclaw2026}; (c) ChemCost
  ~\cite{wu2026chemcost}; (d) MDGym~\cite{kumar2026mdgym}; (e)
  ChemReason-Bench~\cite{zhang2026chemreasonbench}; (f) MADE
  ~\cite{malik2026made}; (g) Corral MD~\cite{alampara2025corral}; (h)
  ScienceAgentBench~\cite{chen2025scienceagentbench}; (i) SciAgentBench
  chemistry~\cite{shen2026sciagentgym}; (j) SciCode~\cite{tian2024scicode};
  (k) CORE-Bench~\cite{siegel2024corebench}; and (l) SciAgentArena molecule
  optimization~\cite{liu2026sciagentarena}. Asterisks mark overall results used
  where no numeric molecular subgroup is published. In (l), darker cells denote
  higher task scores on the source's 0--1 scale and hatching denotes incompatible
  agent--task combinations. Axes retain source metrics and are comparable only
  within panels.}
  \Description{Twelve compact small-multiple panels provide one reported result
  slice for every molecular or scientific-agent benchmark in Table 2. The final
  panel is a ten-task molecule-optimization heatmap with hatched incompatible
  agent--task combinations. Three panels are explicitly marked as overall
  results because their sources do not publish numeric molecular subgroups.}
  \label{fig:benchmark_results_compact}
\end{figure*}

Because benchmark-specific scores are not on a common scale, Figure~\ref{fig:benchmark_results_compact} provides one compact, source-faithful result slice for each benchmark in
Table~\ref{tab:agent_benchmark_landscape}, rather than a cross-benchmark aggregate. 
Molecular-dynamics or chemistry subsets are used when a primary source tabulates them. ScienceAgentBench and SciCode include computational chemistry or chemistry problems but do not tabulate model-level chemistry scores, and CORE-Bench has no molecular split; their panels are therefore explicitly marked as overall results. Each panel retains its source metric and must be interpreted locally.

Figure~\ref{fig:benchmark_results_compact} supports four observations when the scores are read together with each benchmark's design, rather than as a shared leaderboard.

\noindent\textbf{Insight 1: benchmark scores answer different scientific questions.} ChemReason-Bench and ChemCost evaluate bounded procedural or cost reasoning against specified targets~\cite{zhang2026chemreasonbench,wu2026chemcost}.
MolViBench, MDGym, Corral, and SciAgentBench instead require executable tool sequences, making routing, intermediate validity, and recovery part of the evaluated object~\cite{li2026molvibench,kumar2026mdgym,alampara2025corral,shen2026sciagentgym}.
ScienceAgentBench, SciCode, and CORE-Bench extend the horizon to research code or reproducibility artifacts, whereas MADE and the SciAgentArena optimization slice evaluate feedback-guided search under an oracle budget~\cite{chen2025scienceagentbench,tian2024scicode,siegel2024corebench,malik2026made,liu2026sciagentarena}.
Thus, superficially similar percentages can denote answer accuracy, executable completion, reproduction success, or search efficiency.

\noindent\textbf{Insight 2: outcomes are configuration-sensitive, but more tools are not uniformly better.} In the reported MolBench slice, MolClaw-CC reaches 81.1\% accuracy versus 51.4\% for vanilla agents and a 45.1\% standalone-LLM mean. For matched ChemCost backbones, ReAct with tools is associated with a CTA@25 change from 2.5--4.1\% to 34.6--50.6\%~\cite{molclaw2026,wu2026chemcost}. 
Yet SciAgentBench's OtherTools ablation scores 21.0\% versus 28.6\% for the base configuration, and Corral changes the relative ordering of ReAct and tool calling across backbones and single versus chained tasks~\cite{shen2026sciagentgym,alampara2025corral}. Framework claims therefore need matched backbone, controller, tool-set, and task conditions.

\noindent\textbf{Insight 3: longer or harder execution exposes a reliability gap.} SciCode reports 21.2--28.5\% subproblem Pass@1 but only 1.5--7.7\% on complete main problems~\cite{tian2024scicode}. MDGym full success falls from at most 21\% on easy tasks to at most 4\% on hard tasks, while CORE-Bench drops from 60.0\% to 21.5\% for GPT-4o and from 44.4\% to 16.3\% for GPT-4o-mini ~\cite{kumar2026mdgym,siegel2024corebench}.
Across these distinct designs, the pattern is consistent with errors accumulating across specification, execution, diagnosis, and repair. Partial credit is informative, but should be paired with end-to-end success and results stratified by horizon or difficulty.

\noindent\textbf{Insight 4: the metric must expose efficiency and coverage, not only success.} MADE reports acceleration over a random baseline under a fixed oracle budget, so it measures how efficiently feedback redirects discovery rather than whether a single answer is correct~\cite{malik2026made}.
ScienceAgentBench separately reports partial/full success and expert-knowledge conditions, revealing that added context does not remove the execution gap in the displayed settings~\cite{chen2025scienceagentbench}.
SciAgentArena's heatmap further combines strong, failed, and unsupported agent--task pairs~\cite{liu2026sciagentarena}; a mean over completed cells would therefore hide both specialization and missing coverage.

These characteristics also delimit autonomy claims. L1 can use bounded correctness or fixed-workflow completion, whereas L2 needs evidence that an observation changes a candidate, plan, tool choice, or stopping decision, along with loop completion, recovery, cost, and oracle use. L3 additionally requires repeatable physical execution, measurement-conditioned decisions, human interventions, and safety violations, consistent with multidimensional self-driving-laboratory evaluation~\cite{volk2024performance}. No benchmark in Table~\ref{tab:agent_benchmark_landscape} yet establishes a general L3 standard or the cross-campaign agenda revision required for L4. Accordingly, headline scores should be accompanied by task coverage, intermediate validity, end-to-end success, efficiency, reproducibility, operational domain, and mandatory human gates.

%% file: sections/04autonomy.tex
% Recommended packages:
% \usepackage{amsmath,amssymb}
% \usepackage{booktabs,tabularx}

\section{Evidence for Scientific Autonomy Levels}
\label{sec:autonomy_levels}

Molecular LLM agents differ in their scientific capabilities and in the extent to which they independently control a scientific workflow. Laboratory-autonomy frameworks separate physical process execution from data analysis, interpretation, decision-making, and communication, and they classify systems partly by the decisions that still require a human researcher~\cite{hung2024autonomous,volk2024performance}. We follow this emphasis on decision authority while adapting it to LLM-centered molecular agents.
Implementation features such as planning, memory, reinforcement learning, tool use, or multi-agent coordination remain enabling mechanisms: a multi-agent system may require approval at every consequential step, whereas one controller may complete a narrow experimental episode without such a gate.

As shown in Figure~\ref{fig:architecture}, we define scientific autonomy according to the \emph{outermost feedback loop that an agent can reliably close without mandatory human intervention}. This is an operational rubric for this survey rather than a claim that the field has converged on one universal scale.
The distinction concerns decision responsibility and the source of feedback, not the number of tools, reasoning steps, model components, or repeated trials.

\paragraph{Scientific workflows as nested feedback loops.} Let $\tau$ denote one scientific workflow episode executed by agent $A$. We consider three nested forms of loop closure: 
\begin{equation}
d\in \{\mathrm{comp},\mathrm{phys},\mathrm{sci}\},
\end{equation}
corresponding respectively to computational, experimental, and scientific-agenda loops.

For each loop type $d$, define
\begin{equation}
C_d(A,\tau)
=
\mathbb{I}
\left[
\mathsf{Loop}_d(A,\tau)
\land
H_d(\tau)=0
\right],
\label{eq:loop_indicator}
\end{equation}
where $\mathsf{Loop}_d(A,\tau)$ indicates that the corresponding feedback loop is completed, $H_d(\tau)$ is the number of mandatory human decision gates inside that loop, and $\mathbb{I}[\cdot]$ is the indicator function.

A mandatory human gate is an intervention without which the ordinary workflow cannot continue. Human monitoring, retrospective inspection, emergency stopping, and specification of initial constraints do not count as mandatory gates when the agent can otherwise proceed. By contrast, required protocol verification before instrument execution is a gate; AutoLabs explicitly reports such a verification step~\cite{panapitiya2026autolabs}.

The three indicators are defined as follows.

\begin{itemize}
    \item
    $C_{\mathrm{comp}}=1$ if the agent uses tool or environment evidence to revise a scientifically consequential state or decision, such as a candidate, hypothesis, plan, tool choice, or stopping rule, without an intermediate human decision. When no physical experiment is controlled, this evidence must come from a digital or simulated environment. Executing a fixed pipeline or repairing only syntax does not satisfy this criterion.

    \item
    $C_{\mathrm{phys}}=1$ if $C_{\mathrm{comp}}=1$ and the agent additionally designs or adapts and executes a physical experiment and incorporates execution or measurement feedback into recovery, state, interpretation, or completion without an intermediate human decision. Selecting a subsequent experiment from measurements is a stronger iterative form of physical autonomy, but is not required for $C_{\mathrm{phys}}=1$.

    \item
    $C_{\mathrm{sci}}=1$ if $C_{\mathrm{phys}}=1$ and the agent additionally uses accumulated evidence to formulate, prioritize, and pursue new scientific questions, including the autonomous selection of transitions across different scientific task families.
\end{itemize}

These loop types are hierarchical:
\begin{equation}
C_{\mathrm{sci}}(A,\tau)
\leq
C_{\mathrm{phys}}(A,\tau)
\leq
C_{\mathrm{comp}}(A,\tau).
\label{eq:loop_hierarchy}
\end{equation}
Scientific-agenda autonomy therefore presupposes the ability to manage the experimental and computational processes required to investigate selected questions. Physical autonomy likewise presupposes adaptive planning, analysis, or control. A robot that only replays a fixed protocol does not satisfy either $C_{\mathrm{comp}}$ or $C_{\mathrm{phys}}$. This criterion separates automation of execution from adaptive decision-making, a distinction also made in laboratory-autonomy frameworks~\cite{hung2024autonomous}.

\paragraph{Autonomy-level definition.} Given the nested loop indicators, the
episode-level autonomy of agent $A$ is defined as
\begin{equation}
L(A,\tau)
=
1
+
C_{\mathrm{comp}}(A,\tau)
+
C_{\mathrm{phys}}(A,\tau)
+
C_{\mathrm{sci}}(A,\tau).
\label{eq:autonomy_level}
\end{equation}
Because of Eq.~\eqref{eq:loop_hierarchy}, $L(A,\tau)\in\{1,2,3,4\}$.

This produces the following four levels.

\begin{table}[t]
\centering
\footnotesize
\setlength{\tabcolsep}{4pt}
\renewcommand{\arraystretch}{1.12}
\caption{Levels of scientific autonomy for molecular LLM agents.}
\label{tab:autonomy_levels}
\begin{tabularx}{\columnwidth}{@{}c l X@{}}
\toprule
\textbf{Level} &
\textbf{Closed loop} &
\textbf{Operational criterion} \\
\midrule

L1 & None & The agent retrieves information, invokes tools, provides recommendations, or executes a fixed workflow, but evidence does not autonomously revise a consequential scientific decision. \\

L2 & Adaptive computational & Given a computational objective, the agent uses digital evidence to revise a candidate, hypothesis, plan, tool choice, or stopping decision within a digital or simulated environment. \\

L3 & Physical workflow & Given a high-level experimental objective, the agent designs and executes a physical experiment and incorporates execution or measurement feedback without a mandatory human gate during ordinary operation. \\

L4 & Scientific agenda & The agent formulates and prioritizes new scientific questions, selects suitable computational and experimental task families, and updates its research agenda from accumulated evidence. \\

\bottomrule
\end{tabularx}
\end{table}

The four levels may therefore be summarized as follows: L1 systems remain assistive or fixed; L2 systems adapt a computational scientific state from digital evidence; L3 systems autonomously complete a feedback-aware physical workflow; and L4 systems additionally control the evolution of the scientific agenda.

\begin{table*}[t]
\centering
\caption{Representative molecular-agent systems classified by the strongest feedback loop demonstrated in their reported evaluation. The examples are illustrative rather than exhaustive; level is assigned from evidence-conditioned behavior, not from the number of tools or agents.}
\label{tab:representative_agent_levels}
\scriptsize
\setlength{\tabcolsep}{3.5pt}
\renewcommand{\arraystretch}{1.10}
\begin{tabularx}{\textwidth}{@{}>{\raggedright\arraybackslash}p{2.45cm}>{\raggedright\arraybackslash}p{3.05cm}>{\raggedright\arraybackslash}X>{\raggedright\arraybackslash}p{2.55cm}@{}}
\toprule
\textbf{System} & \textbf{Scientific setting} & \textbf{Strongest demonstrated feedback or action} & \textbf{Assigned level} \\
\midrule
ChemAgent~\cite{tang2025chemagent} & Chemical reasoning & Updates task memory to improve bounded reasoning, without closing an external scientific feedback loop & L1: assistive memory \\
CACTUS~\cite{mcnaughton2024cactus} & Chemistry tool use & Executes a predefined tool-mediated problem-solving workflow & L1: fixed workflow \\
Chemist-X~\cite{chen2023chemist} & Reaction conditions & Retrieves evidence and recommends conditions; downstream experimental revision remains external & L1: recommendation \\
ChemCrow~\cite{m2024augmenting} & General chemistry & Uses computational chemistry tools and their outputs to revise a tool-mediated solution trajectory & L2: computational loop \\
ChatDrug~\cite{liu2024conversational} & Molecular editing & Uses retrieval and domain feedback to revise subsequent molecular edits & L2: feedback-driven edit \\
ChemReasoner~\cite{sprueill2024chemreasoner} & Catalyst and molecular search & Uses quantum-chemical rewards to redirect search over candidate reasoning paths & L2: simulation feedback \\
MDCrow~\cite{campbell2026mdcrow} & Molecular dynamics & Uses execution and analysis feedback to diagnose and repair computational MD workflows & L2: workflow adaptation \\
TRACE~\cite{li2026molecular} & Lead optimization & Reuses tool trajectories and changes candidates or tool plans from computed evidence & L2: optimization loop \\
Coscientist~\cite{boiko2023autonomous} & Chemical experimentation & Plans and executes physical experiments and incorporates execution feedback during ordinary operation & L3: physical workflow \\
LLM-RDF~\cite{ruan2024automatic} & Robotic chemistry & Converts plans into robotic execution and interprets experimental readouts & L3: experiment feedback \\
ORGANA~\cite{darvish2025organa} & Automated laboratory & Generates procedures, executes them physically, and uses observed results in completion decisions & L3: experiment feedback \\
AutoLabs~\cite{panapitiya2026autolabs} & Laboratory protocol generation & Self-corrects hardware-ready procedures, followed by required human verification before execution & L2: computational workflow \\
\textit{No surveyed system} & Cross-campaign discovery & No evaluated system demonstrates evidence-conditioned formulation and revision of a scientific agenda across task families & L4: evidence gap \\
\bottomrule
\end{tabularx}
\end{table*}

\paragraph{Clarifying the level boundaries.} The number of tool calls or agents is not an autonomy criterion. A workflow involving many tools remains at L1 when its sequence is fixed or when outputs do not change a later scientific decision. A relatively simple optimization workflow may qualify as L2 if computational evidence changes the candidate, plan, tool choice, or stopping rule. ReAct-style task completion, format repair, and review-only debate do not qualify by themselves.

Physical execution alone is also insufficient for L3. A robotic platform that replays a fixed human-written protocol performs automation. L3 requires the agent to formulate or adapt the experimental procedure and to use physical feedback in recovery, interpretation, or completion.

The statement that L3 \emph{does not require a second experiment} refers only to the number of physical trials, not to the absence of feedback. A single episode can qualify when an observation changes what the agent does or concludes inside that episode. For example, Coscientist reads UV--Vis spectra to identify an initially unknown physical state and, in a separate integrated experiment, consults hardware documentation and repairs an invalid automation method before successful execution~\cite{boiko2023autonomous}. ORGANA likewise uses visual feedback to guide long-horizon physical plans~\cite{darvish2025organa}. Neither example needs a second synthesis to demonstrate feedback-aware physical control. When a measured yield or material property selects another experiment, as in LLM-RDF and robotic
ChemAgents~\cite{ruan2024automatic,song2025multiagent}, we label the evidence \emph{iterative L3}. This distinction is specific to our physical-workflow rubric; SDL taxonomies often reserve their stronger closed-loop categories for systems that also automate experiment selection across repeated trials~\cite{volk2024performance}.

Generating candidate hypotheses is also insufficient for L4. The agent must evaluate candidate questions, select one to pursue, allocate computational or experimental actions, and revise its research direction from the evidence.
Transitions between task families must be evidence-conditioned rather than fixed entirely in advance.

\paragraph{Domain-conditioned autonomy.} An autonomy claim is meaningful only relative to the domain in which the system operates; existing laboratory frameworks likewise caution that a level does not by itself define the scope of the research being categorized~\cite{hung2024autonomous}. Let $\mathcal{D}$ denote a scientific operational domain specifying supported tasks, tools, environments, instruments, resources, and safety constraints.

The system-level autonomy within $\mathcal{D}$ can be defined as
\begin{equation}
L_{\mathcal{D}}(A)
=
\max
\left\{
\ell:
\Pr_{\tau\sim\mathcal{D}}
\left[
L(A,\tau)\geq\ell
\right]
\geq\rho
\right\},
\label{eq:domain_level}
\end{equation}
where $\rho$ is a predefined reliability threshold.

Safety can be imposed as a separate evaluation constraint:
\begin{equation}
\Pr_{\tau\sim\mathcal{D}}
\left[
U(A,\tau)=1
\right]
\leq\delta,
\label{eq:safety_constraint}
\end{equation}
where $U(A,\tau)$ indicates an unsafe or invalid episode and $\delta$ denotes the maximum acceptable violation rate.

Separating Eqs.~\eqref{eq:domain_level} and \eqref{eq:safety_constraint} is important because autonomy and competence are distinct. A system may be authorized to complete an L3 workflow but do so unreliably, whereas a highly accurate molecular predictor may remain at L1 because it neither determines nor executes subsequent actions.

\paragraph{Complementary continuous measures.} The discrete level can be accompanied by continuous measures of human involvement and loop reliability. This follows proposals to report autonomy alongside lifetime, throughput, precision, material use, accessible parameter space, and optimization performance~\cite{volk2024performance}. For example, the mandatory human-intervention rate may be reported as
\begin{equation}
R_{\mathrm{H}}
=
\frac{N_{\mathrm{gate}}}
     {N_{\mathrm{decision}}},
\end{equation}
where $N_{\mathrm{gate}}$ is the number of mandatory human gates and $N_{\mathrm{decision}}$ is the number of consequential workflow decisions.

For loop type $d$, the empirical loop-completion rate is
\begin{equation}
R_d
=
\frac{1}{N}
\sum_{i=1}^{N}
C_d(A,\tau_i).
\end{equation}
These measures distinguish systems that nominally belong to the same autonomy level but differ substantially in robustness or dependence on human intervention.

\paragraph{Relationship to enabling technologies.} Planning, memory, reinforcement learning, and multi-agent coordination are orthogonal to the autonomy levels. They are enabling mechanisms whose roles change across levels rather than level-defining properties.

At L1, memory may support information retrieval and personalized assistance.
At L2, it may preserve state across long computational workflows and support recovery from failed tool calls. At L3, it may maintain experimental context across physical iterations. At L4, it may organize evidence accumulated over multiple projects and support long-term research-agenda formation.

The same principle applies to reinforcement learning and multi-agent coordination. Their contribution to autonomy should be assessed by whether they enable the system to close a broader feedback loop, rather than by their mere presence in the architecture.

%% file: sections/05safety_challenges.tex
\section{Safety and Challenges}

\begin{figure}[t]
\centering
\includegraphics[width=1.0\columnwidth]{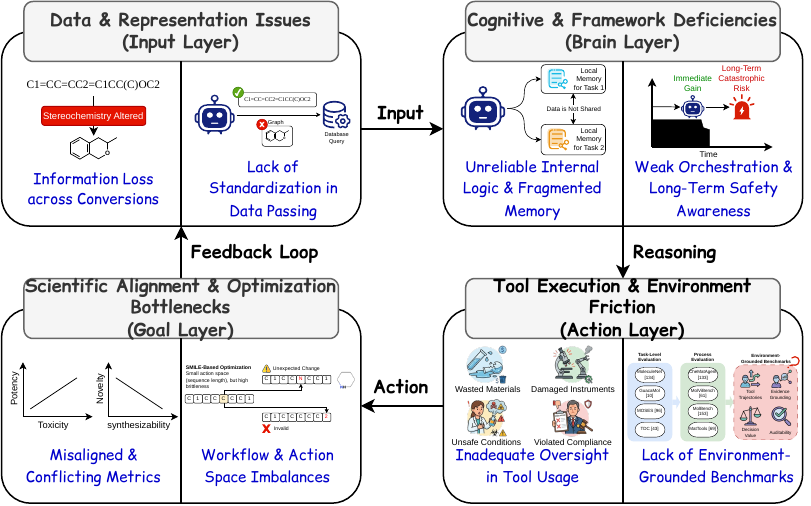}
\caption{Challenges and bottlenecks in the lifecycle of a molecular agent. The diagram maps vulnerabilities across perception, controller design, tool and environment interaction, and optimization; arrows indicate how an upstream error can propagate into more consequential system-level behavior as action authority increases.}
\Description{A four-layer molecular-agent lifecycle showing risks in molecular perception, agent control, tool and environment interaction, and optimization. Arrows indicate that failures can propagate across layers and grow more consequential as the system gains action authority.}
\label{fig:challenges}
\end{figure}

The architectural components in Section~\ref{sec:architecture} expose distinct
but coupled failure modes. Perception can corrupt molecular identity, the agent
core can select an unsupported action, a tool can return unreliable evidence,
and optimization can amplify weaknesses in any of the preceding components. In this section, we
provide an overview of the safety concerns and then organize the open challenges along the same four-part architecture which can be summarized as Figure~\ref{fig:challenges}.

\subsection{Safety and Governance Boundaries}

The risk posed by a molecular agent is determined by its operational authority, the reversibility of its actions, the scope of substances and equipment it can affect, and the consequences of an erroneous decision. The
self-driving-laboratory safety literature accordingly treats safety as a
property of interactions among software, hardware, materials, people, and operating
procedures rather than of the language model alone~\cite{leong2025safe}.
For instance, read-only retrieval and identity checks can be assigned to a control
surface narrower than purchasing, synthesis planning, or instrument control. Furthermore, the same tool may also require different permissions across operational domains.

The safety concerns raised by molecular discovery agents can be classified into three broad categories: intrinsic system limitations, extrinsic human-induced threats, and physical-world operational risks.

At the intrinsic level, these agents rely on reasoning, planning, and memory modules that are prone to instability or insufficiency, which can lead to hallucinations or erroneous predictions of molecular properties. Such inaccuracies introduced during early-stage information retrieval can cascade into catastrophic failures through synthesis planning and robotic execution~\cite{he2023controlriskpotentialmisuse,cao2026mozi}. When such erroneous information guides experimental workflows, it can resulting in resource waste, generation of hazardous byproducts, and potential laboratory incidents. Moreover, large language models are known to struggle with long-horizon planning and complex reasoning tasks, impairing the agent’s ability to anticipate downstream consequences or recognize critical safety checkpoints in multi-step synthetic routes~\cite{Tang2024Risks}.

On the human–agent interface, extrinsic safety is predominantly challenged by the diversity of user intent. Instructions may range from benign requests to deliberately malicious prompts. Adversarial inputs and jailbreak techniques can subvert the agent’s safety alignment, compelling it to propose synthetic pathways for hazardous compounds. Dual use is a concrete molecular design concern, the same capabilities that enable beneficial drug discovery can be repurposed to design highly toxic compounds, controlled substances, or chemical weapons~\cite{urbina2022dualuse}.

However, the most consequential safety challenge emerges when molecular discovery agents are integrated with automated laboratory platforms. In such scenario, digital decision errors translate directly into physical consequences. A single erroneous command affecting reagent sequencing, temperature control, or hazardous material handling can trigger chemical spills, fires, explosions, or personal injury~\cite{cao2026mozi}. This autonomy risk fundamentally distinguishes molecular discovery agents from general-purpose large language models. Their capacity for direct material manipulation demands rigorous risk control mechanisms, sustained human oversight, and a safety evaluation that is specifically tailored to autonomous molecule discovery.

For the systems surveyed here, we recommend a governance contract that records
the verified molecular identity, allowed tools and resources, approval gates,
stopping and abort conditions, and an auditable trace of inputs, intermediate
states, tool versions, outputs, and overrides. This follows the lifecycle view
of the NIST AI Risk Management Framework, whose core functions are to govern,
map, measure, and manage risk~\cite{tabassi2023airmf}. Chemistry-specific
implementations include safety tools in ChemCrow~\cite{m2024augmenting},
governed skills in Mozi~\cite{cao2026mozi}, and bounded laboratory interfaces
in Coscientist~\cite{boiko2023autonomous}. Evaluation metrics should then capture not only task success but also invalid or unsafe episodes, blocked or escalated actions, budget violations, recovery events, and human interventions. These are proposed reporting
requirements for this survey, not claims that any single control stack is sufficient
for every laboratory.

\input{sections/0501Perception_Challenges}

\input{sections/0502Agent_Design_Challenges}

\input{sections/0503Tool_Challenges}

\input{sections/0504Optimization_Challenges}

%% file: sections/0501Perception_Challenges.tex
\subsection{Challenges in Molecular Perception}

Molecular perception must preserve chemical identity while translating among representations selected for different tasks. This requirement becomes harder as agents move from human-checked L1 interactions to long, autonomous, and multimodal trajectories at L2 and above.

\paragraph{Representation fidelity across conversions.}
No single substrate exposes every relevant molecular property. SMILES and SELFIES are convenient for generation, graphs expose topology, 3D structures support physical reasoning, and images or spectra connect the agent to literature and experiments. Routing among these substrates can silently alter stereochemistry, protonation, tautomeric state, atom mapping, conformers, or measurement context. Structured formats such as MolJSON and MoleCode make connectivity more explicit~\cite{runcie2026moljson,liu2026molecode}, but future agents still need round-trip validation, invariant checks, and persistent identity tracking at every representation boundary.

\paragraph{Adaptive representation and token granularity.}
Representation choice should depend on the next scientific action rather than on a fixed input format. A compact string may be sufficient for database lookup, whereas exact editing, docking, or spectroscopy may require graphs, coordinates, or multimodal state. Tokenization further determines whether edits correspond to characters, atoms, fragments, or structure-aware units.
The open problem is to learn when the current representation is inadequate and to switch substrates without losing information, exceeding the context budget, or obscuring the action from human review.

\paragraph{Uncertainty-aware feedback perception.}
Tool outputs and experimental observations are not self-interpreting facts.
Docking scores, property predictions, spectra, and assay measurements depend on model assumptions, units, conditions, and domains of validity. An agent must retain these qualifications when converting an observation into its next state. Otherwise, an ambiguous experimental signal or an out-of-domain prediction can be collapsed into false certainty and propagated through later decisions. Perception benchmarks should therefore test identity preservation, unit and condition tracking, uncertainty retention, and recovery from conflicting multimodal evidence together with structure parsing accuracy.

%% file: sections/0502Agent_Design_Challenges.tex
\subsection{Challenges in the Agent Framework}

Although molecular agent design has progressed from tool-using dialogue to
workflow control, reliable autonomous discovery remains unresolved.
Open problems concern verifiability, memory, orchestration, evaluation, and
safety.

\paragraph{Verifiable and molecule-native reasoning.}
Current agents can generate structured tool calls, executable code, protocols, and action traces, but they do not yet provide machine-checked chemical proofs or formal constraint certificates~\cite{li2026chemical}.
ChemActor represents synthesis procedures as structured chemical actions~\cite{zhang2025chemactor}, but formal verification of molecule edits, synthesis constraints, safety rules, and protocol preconditions remains open.
The reasoning substrate is also unresolved.
Text and SMILES are interpretable but lossy, while graph, 3D, and latent states better preserve molecular structure but are harder to audit.
MolLingo~\cite{nguyen2026mollingo} and LatentChem~\cite{ye2026latentchem} illustrate this tension between molecule-native expressiveness and interpretability.

\paragraph{Persistent memory and reliable reflection.}
Most memory remains local to one task, session, or campaign.
ChemAgent~\cite{tang2025chemagent} uses self-updating memories,
TRACE~\cite{li2026molecular} stores tool trajectories, and Augmented
Memory~\cite{guo2024augmented} uses experience replay. L4 discovery would
require cross-task scientific memory for failures, provenance, uncertainty,
design rules, and reusable skills.
Reflection faces a similar limitation.
Systems such as ChatDrug~\cite{liu2024conversational}, AgentDrug~\cite{le2024agentdrug}, Probe-Before-You-Edit~\cite{yang2026probe}, and MT-Mol~\cite{kim2025mt} rely on external feedback, but high-autonomy agents often face delayed, noisy, or missing ground truth.
Future evaluation of reflection must therefore consider calibration, stopping rules, and when the agent should defer to simulation, experiment, or human review.

\paragraph{Adaptive orchestration and evaluation.}
More tools, memory, and agents are not always better.
Tooling-or-Not-Tooling shows that tool use can help some chemistry tasks while hurting others~\cite{yu2025tooling}.
Orchestration should be adaptive: the controller must decide when to answer directly, retrieve evidence, call tools, coordinate specialists, or request human approval.
Evaluation is also underdeveloped.
Most benchmarks measure final task scores rather than controller behavior, such as reasoning traces, tool routing, planning under budget, memory reuse, reflection after failure, and multi-agent coordination.
ChemLabs~\cite{xu2025chemlabs} and Tooling-or-Not-Tooling~\cite{yu2025tooling} are useful steps, but environment-grounded molecular-agent benchmarks remain limited.

\paragraph{Safety and governance.}
The controller-level problem is to enforce the boundaries defined above at the
point of decision making. A final text filter cannot validate molecular identity,
instrument state, or whether a requested action lies inside the approved
domain. Dual-use molecular generation~\cite{urbina2022dualuse} and the expanding
scope of self-driving laboratories~\cite{leong2025safe} motivate
structure-aware screening, permissioned execution, traceable decisions, and
escalation when evidence or authority is insufficient. The open research
question is how to assess these controls without relying solely on refusal rate
as the safety metric.

%% file: sections/0503Tool_Challenges.tex
\subsection{Challenges in Molecular Toolboxes}

Tools ground molecular agents in external evidence and executable actions, but
they also define the operational boundary within which an agent can fail. A
reliable toolbox must make capabilities, assumptions, costs, and hazards
visible to the controller.

\paragraph{Semantic interoperability and state passing.}
Chemistry tools use heterogeneous identifiers, file formats, units, parameter
conventions, and software environments. Ad hoc wrappers hide these differences
but rarely guarantee that a state produced by one tool is valid input to the
next. Shared, typed interfaces are needed for molecular identity, experimental
conditions, uncertainty, provenance, and pre- and postconditions. General
scientific tool infrastructures such as ToolUniverse~\cite{gao2025tooluniverse}
provide a useful foundation, but molecular workflows additionally require
chemistry-aware schemas and validation across databases, simulators, synthesis
planners, instruments, and robots.

\paragraph{Reliability, calibration, and failure recovery.}
Tool access does not guarantee trustworthy grounding. Databases can disagree,
predictors can be out of domain, simulations can fail to converge, and
laboratory interfaces can return incomplete observations. Agents must detect
invalid inputs, timeouts, numerical failures, inconsistent outputs, and low
confidence before using a result to revise the plan. Because tool augmentation
can either help or hurt depending on the task~\cite{yu2025tooling}, evaluation
should measure tool selection, input construction, output interpretation,
fallback behavior, and calibration in addition to final-answer accuracy.

\paragraph{Reproducibility, cost, and governed execution.}
Long molecular workflows require versioned tools, recorded parameters,
environment metadata, and traceable state transitions so that a result can be
reproduced rather than merely narrated. The controller must also trade off
information value against latency, compute, assay cost, and instrument access.
For synthesis or laboratory action, these resource policies must connect to the
run-level permissions, abort conditions, and audit record defined in the safety
governance subsection. The toolbox-specific challenge is implementing those
controls across heterogeneous software and hardware interfaces while keeping
execution interruptible and confined to the approved operational domain.

%% file: sections/0504Optimization_Challenges.tex
\subsection{Challenges in Molecular Optimization}

Molecular optimization tests whether an LLM-based molecular agent can improve candidates under scientific constraints, beyond generating valid structures.
Agents must search a vast chemical space under noisy feedback, incomplete constraints, and conflicting objectives.
Open problems lie in reward design, multi-objective search, action control, and workflow-level feedback.

\paragraph{Proxy objectives and reward hacking.}
Most optimization systems rely on surrogate objectives, such as QED, penalized LogP, docking scores, predicted binding affinity, synthetic accessibility, and ADMET predictors.
Early reinforcement-learning methods showed that molecular generators can be guided by property rewards~\cite{olivecrona2017molecular,popova2018deep}, while REINVENT4 supports richer multi-component scoring functions~\cite{loeffler2024reinvent4}.
These scores provide fast feedback, but they remain proxies for the real scientific goal.
An agent may exploit the evaluator and produce molecules that score well but are unstable, toxic, difficult to synthesize, or outside the predictor's reliable domain.
The challenge is to design rewards that consider validity, synthesizability, uncertainty, diversity, and experimental plausibility rather than optimize a single scalar score.

\paragraph{Multi-objective trade-offs and local optima.}
Real molecular design usually involves conflicting objectives, where improving one property may weaken another.
For example, higher potency may increase toxicity, while higher novelty may reduce synthesizability.
DrugEx v2~\cite{liu2021drugex}, QADD~\cite{qadd2023}, and MARS~\cite{xie2021mars} reflect the move from single-objective optimization toward Pareto ranking, iterative multi-objective search, and graph-based editing.
For agent-based optimization, premature exploitation is another concern.
Once an agent finds a high-scoring scaffold, it may keep making small local edits around it rather than explore new chemical regions.
Exploration-oriented methods such as Mol-AIR~\cite{molair2024} and Augmented Hill-Climb~\cite{thomas2022augmented} address part of this problem through novelty, coverage, or more efficient search.
Future agents need better mechanisms for Pareto-aware optimization, diversity preservation, and control of exploration and exploitation.

\paragraph{Search efficiency and action-space control.}
Chemical space is too large for blind trial and error, especially when evaluation requires docking, molecular dynamics, DFT, retrosynthesis analysis, or laboratory validation.
SMILES-based optimization is easy to combine with sequence models~\cite{olivecrona2017molecular,popova2018deep,loeffler2024reinvent4}, but small token-level edits can invalidate a molecule or change it unexpectedly.
Graph-based methods such as GCPN~\cite{you2018graph} and MolDQN~\cite{zhou2019optimization} make actions more chemically explicit by operating on atoms, bonds, and graph edits.
However, they also introduce larger and more constrained action spaces.
The challenge is to define actions that are expressive enough for discovery, constrained enough to avoid invalid chemistry, and efficient enough for multi-round optimization.

\paragraph{Workflow-level feedback and credit assignment.}
Molecular optimization increasingly operates as a multi-step workflow rather than a single generation step.
LLM-guided and tool-using systems such as ReMol~\cite{wang2025remol}, ChemCrow~\cite{m2024augmenting}, ChemCRAFT~\cite{li2026agentic}, and MolClaw~\cite{molclaw2026} show this shift from molecular generators toward broader agentic pipelines.
When optimization fails, the cause may lie in generation, tool selection, reward modeling, planning, or molecular format conversion.
This creates a workflow-level credit assignment problem.
Evaluation should therefore measure final molecular scores together with multi-round improvement, tool-use efficiency, failure recovery, provenance, and robustness to unreliable feedback.

%% file: sections/06conclusion.tex
\section{Future Directions}

\input{sections/0602Future_Work}

\section{Conclusion}

This survey developed a conceptual framework for molecular LLM agents from two complementary perspectives. The architectural view connects molecular representation and perception, an LLM-centered agent framework, domain-specific toolboxes, and learning and optimization. The scientific-autonomy view classifies agents by the outermost feedback loop they can reliably close without mandatory human intervention, from L1 assistive or fixed workflows to L4 scientific-agenda agents. Together, these views connect system design with demonstrated decision authority across computational and physical molecular workflows.

Our analysis indicates that planning, tool use, stronger language models, or multi-agent coordination alone do not establish scientific autonomy. Progress also requires chemically faithful perception, grounded and verifiable tool use, feedback-aware decision making, persistent provenance, realistic optimization, and safety within an explicit operational domain. Future evaluation should therefore consider task success together with loop completion, uncertainty, cost, reproducibility, human involvement, and safety, helping molecular agents advance from useful assistants toward trustworthy partners in molecular discovery.

%% file: sections/0602Future_Work.tex
The open problems above become more useful when expressed as testable milestones rather than as another list of missing capabilities. We propose a progression in which each stage produces artifacts and measurements needed to justify the next expansion of authority.

\paragraph{Milestone 1: a verifiable molecular-state contract.} Near-term systems should publish typed schemas for molecular identity, structure, conditions, units, uncertainty, provenance, and permitted edits, together with round-trip conversion tests across strings, graphs, coordinates, images, spectra, and experimental records. MolJSON, MoleCode, and MolLingo~\cite{runcie2026moljson,liu2026molecode,nguyen2026mollingo} illustrate current attempts to make molecular structure more explicit or molecule-native for LLM reasoning. The milestone is reached when declared conversions preserve identity and required metadata on a public test suite, invalid states are rejected before tool execution, and every downstream result can be traced to its input representation. This is a concrete prerequisite for attributing a later failure to the controller rather than to silent state corruption.

\paragraph{Milestone 2: reproducible and governed L2 loops.} The next target is not a larger tool catalogue but a computational controller whose trajectories can be replayed and compared under the same evidence and resource budget. It should record tool versions, intermediate states, uncertainty, stopping decisions, and the reason for each escalation. Persistent memory can then be evaluated by withholding prior trajectories and measuring whether validated experience improves a new run without propagating stale or incorrect evidence. MolViBench and SciAgentBench expose multi-step tool execution~\cite{li2026molvibench,shen2026sciagentgym}, while MADE supplies a budgeted closed-loop discovery setting~\cite{malik2026made}. A credible L2 milestone should report loop completion, improvement over rounds, recovery, cost, reproducibility, and mandatory human gates against fixed-workflow and no-memory baselines.

\paragraph{Milestone 3: bounded, safety-evaluated physical episodes.}
An L2 controller should progress to L3 only on a declared experimental domain with validated protocols, instrument limits, permissions, abort conditions, and measurement-quality checks. Coscientist and ORGANA provide primary examples of LLM-centered planning connected to physical execution and feedback~\cite{boiko2023autonomous,darvish2025organa}; safety work on self-driving laboratories motivates evaluating the whole software--hardware--material system~\cite{leong2025safe}. The milestone is repeatable completion of the same bounded episode across multiple trials while reporting execution success, measurement validity, recovery, human intervention, unsafe or invalid episodes, and resource use. A successful demonstration should also state what remains outside the operational domain.

\paragraph{Milestone 4: reliable iterative L3 campaigns.}
The stronger experimental target is a loop in which a measurement selects or changes a subsequent physical experiment. LLM-RDF and robotic ChemAgents demonstrate this pattern using measured yield or catalyst performance~\cite{ruan2024automatic,song2025multiagent}. Future work should compare the agent with fixed designs and established acquisition policies under matched budgets, replicate selected measurements, and report calibration, sample efficiency, stopping behavior, and recovery from failed or ambiguous readouts.
This milestone separates an isolated automation success from an experimental policy that learns reliably during a campaign.

\paragraph{Milestone 5: evidence for cross-campaign scientific autonomy.}
L4 should remain an evidence standard rather than an aspirational label. A candidate system would need to preserve evidence across completed campaigns, formulate and prioritize a new question, choose among different computational and experimental task families, allocate resources, and revise the research agenda when results contradict its assumptions. Evaluation would require multiple task families, prospective rather than retrospective trials, independent scientific review, and evidence that the selected question and result are not recoverable from a fixed predefined workflow. Until such studies exist, progress is better described through the measurable L2 and L3 milestones above than through broad claims of autonomous discovery.